\documentclass{article}
\usepackage[numbers]{natbib}

 \usepackage[main, final]{neurips_2025}

\usepackage[utf8]{inputenc} 
\usepackage[T1]{fontenc}    
\usepackage{hyperref}       
\usepackage{url}            
\usepackage{booktabs}       
\usepackage{amsfonts}       
\usepackage{nicefrac}       
\usepackage{microtype}      
\usepackage{xcolor}         

\usepackage{graphicx}       
\usepackage{array} 
\usepackage{multirow}
\usepackage{makecell}
\usepackage{subcaption}
\usepackage{algorithm}
\usepackage{algorithmic}
\usepackage{amssymb}
\usepackage{amsmath}
\usepackage{amsthm}
\usepackage{wrapfig}
\definecolor{lightgreen}{RGB}{105, 164, 76}
\definecolor{seablue}{RGB}{42, 156, 244}

\title{FerretNet: Efficient Synthetic Image Detection via Local Pixel Dependencies}

\author{
Shuqiao Liang \quad Jian Liu \quad Renzhang Chen$\thanks{Corresponding authors.}$ \quad Quanlong Guan$^{*}$ \\
Jinan University\\
\texttt{\{xigua7105, liujian2143\}@gmail.com}, \texttt{\{jnulion, gql\}@jnu.edu.cn} \\
}

\begin{document}

\maketitle

\begin{abstract}
The increasing realism of synthetic images generated by advanced models such as VAEs, GANs, and LDMs poses significant challenges for synthetic image detection.
To address this issue, we explore two artifact types introduced during the generation process: (1) latent distribution deviations and (2) decoding-induced smoothing effects, which manifest as inconsistencies in local textures, edges, and color transitions.
Leveraging local pixel dependencies (LPD) properties rooted in Markov Random Fields, we reconstruct synthetic images using neighboring pixel information to expose disruptions in texture continuity and edge coherence.
Building upon LPD, we propose FerretNet, a lightweight neural network with only 1.1M parameters that delivers efficient and robust synthetic image detection.
Extensive experiments demonstrate that FerretNet—trained exclusively on the 4-class ProGAN dataset—achieves an average accuracy of 97.1\% on an open-world benchmark comprising 22 generative models. Our code and datasets are publicly available at \textcolor{seablue}{\url{https://github.com/xigua7105/FerretNet}}.

\end{abstract}

\section{Introduction}
The field of AI-based image generation has progressed rapidly, driven by the development of powerful generative models such as Variational Autoencoders (VAEs)~\cite{kingma2022autoencodingvariationalbayes}, Generative Adversarial Networks (GANs)~\cite{karras2018progressive,karras2021alias,brock2018large}, and Latent Diffusion Models (LDMs)~\cite{rombach2022ldm,podell2024sdxl,esser2024sd35}.
These models have enabled widespread applications across art, entertainment, and e-commerce, allowing users to effortlessly create realistic and engaging images.
However, the potential misuse of such content has raised ethical concerns and driven extensive research on synthetic image detection~\cite{wang2020cnn,frank2020leveraging,ojha2023towards,liu2024forgery}.

Many existing detection approaches rely heavily on model-specific features, which limit their generalization ability to unseen generative architectures.
For example, Durall et al.~\cite{durall2020watch} observed characteristic frequency artifacts in GAN-generated images.
Although frequency-domain techniques~\cite{wang2020cnn,jeong2022bihpf,jeong2022frepgan} have demonstrated strong performance under known conditions, they often struggle to generalize across different models.
DIRE~\cite{wang2023dire} introduced a diffusion-based detection framework that distinguishes synthetic images by reconstructing them through a diffusion model, a capability that fails with real images.
However, this method performs poorly when applied to GAN-generated content.

To address the generalization challenge, Ojha et al.~\cite{ojha2023towards} explored the utilization of pre-trained models, employed frozen backbone for image encoding, providing universal representations from pre-training, followed by a linear classifier. 
FatFormer~\cite{liu2024forgery} introduced an Adaptor to CLIP~\cite{radford2021learning} to enhance the pre-trained model's ability to learn artifacts. 
While these methods achieved encouraging results, they are constrained by large parameter counts or low computational efficiency.

To overcome the dual challenges of limited generalization and computational inefficiency in synthetic image detection, we conduct a comprehensive analysis of artifact patterns shared across GAN-, VAE-, and LDM-based generative models. Our analysis reveals that visual anomalies-such as unnatural textures, geometric distortions, and poor object-background integration—primarily originate from two sources:
(1) distributional shifts in the latent variable $z$, and (2) over-smoothing and color discontinuities introduced during the decoding process.

Based on these insights and grounded in the theory of Markov Random Fields, we introduce a pixel-level artifact representation that captures local pixel dependencies (LPD) through median-based reconstruction.
We further propose FerretNet, a lightweight detector incorporating depthwise separable and dilated convolutions to achieve a balance between computational efficiency and representational power.

Contributions of this work are as follows:
\begin{itemize}
\item We propose a novel approach that leverages Markov Random Fields and median-based statistics to capture local pixel dependencies for detecting artifacts and anomalies in synthetic images.
\item We present Synthetic-Pop, a 60K-image benchmark for evaluating detection models against high-fidelity generators, containing 30K synthetic images from six models and 30K real images from COCO~\cite{lin2014microsoft} and LAION-Aesthetics V2 (6.5+)~\cite{LaionAesthetics}. See Appendix~\ref{sec: details of dataset} for more details.
\item We introduce FerretNet, a lightweight model with only 1.1 million parameters, which achieves 97.1\% accuracy on synthetic image detection across 22 generative models, while maintaining low computational overhead.
\end{itemize}

\section{Related Work}

We categorize existing synthetic image detection methods into two main paradigms: pixel-based and frequency-based approaches.

\subsection{Pixel-based Synthetic Image Detection}

Wang et al.~\cite{wang2020cnn} trained a classifier on images generated by a single model to detect fake images across various architectures and datasets, addressing cross-model generalization via data augmentation and diverse training samples.
Shi et al.~\cite{shi2023discrepancy} proposed a difference-guided reconstruction learning framework that exploits discrepancies between real and synthetic images to enhance detection accuracy.
Ojha et al.~\cite{ojha2023towards} tackled the generalization problem to unseen generative models by leveraging a feature space not explicitly trained for real/fake discrimination, employing nearest-neighbor and linear probing strategies.
He et al.~\cite{he2021beyond} introduced a super-resolution-based re-synthesis technique to reconstruct test images and extract residual or layered artifact features, thereby reducing reliance on frequency artifacts.
Tan et al.~\cite{tan2024rethinking} proposed NPR, a method that revisits the upsampling process in generative CNNs by modeling Neighbor Pixel Relations, aiming to improve generalization in deepfake detection.
Liu et al.~\cite{liu2023towards} designed a robust detection framework based on multi-view image completion, which simulates real image distributions and captures frequency-independent features.
FatFormer~\cite{liu2024forgery} presented a forgery-aware adaptive transformer incorporating forgery-specific adapters and language-guided alignment modules to better adapt pre-trained models for synthetic image detection.
CO-SPY~\cite{cheng2025co} leverages a frozen CLIP encoder for semantic features and a VAE-based reconstruction difference for artifacts, integrating them via adaptive fusion for robust synthetic image detection.

\subsection{Frequency-based Synthetic Image Detection}

F3Net~\cite{qian2020thinking} introduced a dual-branch architecture that captures frequency-aware clues for detecting subtle forgery traces, particularly in low-quality and facial imagery.
FrePGAN~\cite{jeong2022frepgan} developed a frequency-level perturbation GAN framework, where a generator-discriminator pair is used to iteratively improve classifier robustness against unseen categories and generative models.
Tan et al.~\cite{tan2023learning} exploited pre-trained CNN gradients to generate generalizable representations of GAN-specific artifacts.
BiHPF~\cite{jeong2022bihpf} amplified frequency-level artifacts via a high-pass filtering approach, achieving improved robustness across diverse image categories, color manipulations, and generative models.
FreqNet~\cite{tan2024frequency} introduced high-frequency representations and frequency-specific convolution layers to enhance detection by focusing on localized high-frequency components, addressing overfitting and poor generalization seen in prior methods.
SAFE~\cite{li2025improving} leverages the high-frequency component of the Discrete Wavelet Transform to extract forensic artifacts and employs data augmentation techniques including ColorJitter, RandomRotation, and a patch-based RandomMask mechanism to improve the model’s generalization and robustness.

\section{Artifacts in Synthetic Image Generation}

This section provides a high-level, intuitive framework to motivate our search for universal artifacts. We establish the general principle that all generative models, despite architectural differences, introduce artifacts. While these artifacts have multiple high-level sources (e.g., latent space, decoding), our work chooses to focus on detecting a powerful and universal \textit{effect}—the disruption of local pixel statistics—which is effectively captured by our LPD method.

\subsection{Image Generation Pipeline}

\begin{wrapfigure}{r}{0.5\textwidth}
\vspace{-1.35em}
\centering
\includegraphics[width=0.48\textwidth]{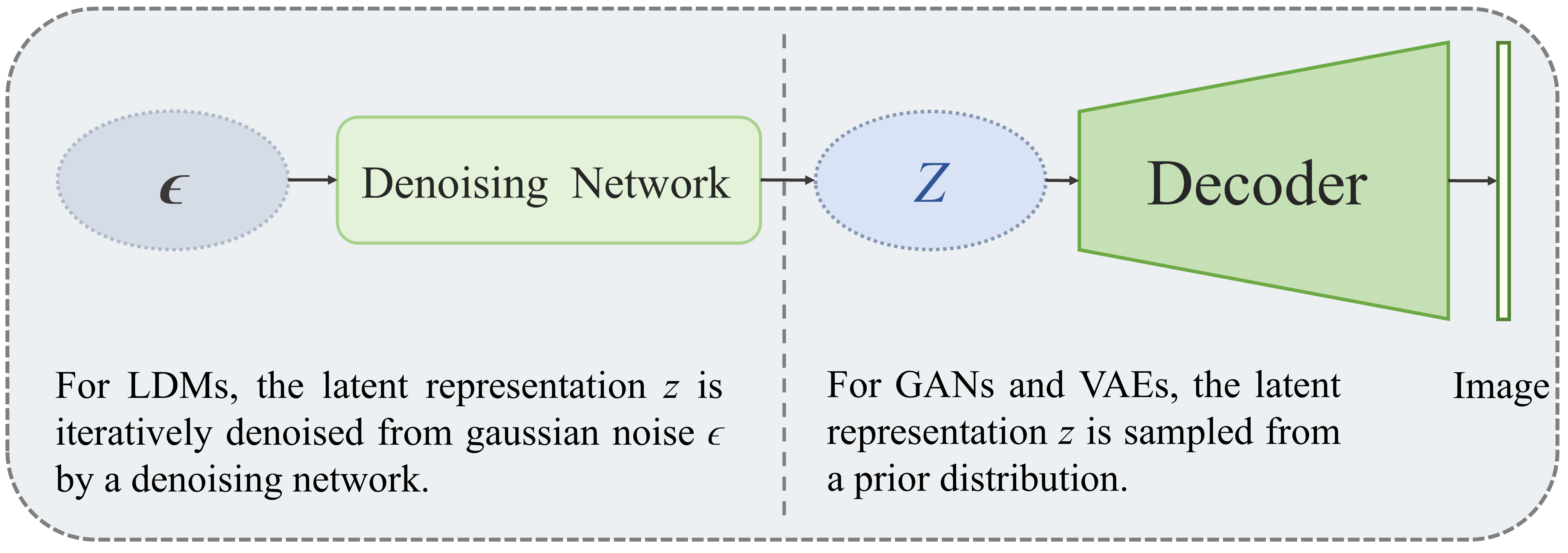}
\caption{The image generation process in VAEs, GANs, and LDMs can be broadly divided into two stages: obtaining the latent variable \( z \), and decoding it into an image.}
\label{fig:Pipeline}
\vspace{-1.0em}
\end{wrapfigure}

Generative models such as VAEs, GANs, and LDMs are widely used for image synthesis. Despite differences in architecture and training objectives, these models share a common two-stage generation pipeline, as illustrated in Figure~\ref{fig:Pipeline}.

\textbf{1. Obtaining the latent variable \(z\):}

In LDMs, the generation process begins with Gaussian noise \( \epsilon \sim \mathcal{N}(0, I) \), which is iteratively denoised into a latent representation \( z \) within the compressed latent space of a pretrained autoencoder, using a denoising network such as U-Net~\cite{ronneberger2015u,rombach2022ldm} or Diffusion Transformer (DiT)~\cite{peebles2023scalable,yao2024fasterdit}. 
In contrast, VAEs and GANs directly sample \(z\) from predefined prior distributions, such as a standard normal distribution \(\mathcal{N}(0, I)\) or a uniform distribution \(U(-1,1)\).

\textbf{2. Decoding \(z\) to generate images:}  
In both VAEs and LDMs, a decoder transforms \(z\) into the final image through a series of convolutional layers with specific kernel sizes and strides. In GANs, the generator plays an analogous role, mapping \(z\) to the image space with the aim of approximating the target data distribution.

While this two-stage framework enables high-fidelity image synthesis, it can also introduce artifacts such as texture irregularities, unnatural transitions, and local detail loss. These artifacts commonly arise from two major sources: (1) deviations in the distribution of the latent variable \(z\), and (2) imperfections introduced during the decoding process.

\subsection{Latent Distribution Deviations}

The quality of synthetic images exhibits significant sensitivity to the distribution of the latent representation \(z\)~\cite{razavi2018preventing,hu2023complexity,DBLP,zhou2024golden}. Ideally, the sampled distribution \(Q(z)\) should match the prior distribution \(P(z)\) assumed or learned during training. However, in practice, factors such as data imbalance or insufficient training can lead to a mismatch between \(Q(z)\) and \(P(z)\). This discrepancy can be quantified using the Kullback–Leibler (KL) divergence:
\begin{align}
D_{\mathrm{KL}}(Q(z) \| P(z)) = \int Q(z) \log \frac{Q(z)}{P(z)} \, dz > \delta,
\end{align}
where \(\delta\) denotes an acceptable divergence threshold. When this threshold is exceeded, the resulting images are prone to visible artifacts, including texture inconsistencies and the loss of fine structural details. For example, in GANs, if the latent space is poorly aligned with the true data distribution, the generator may fail to reproduce realistic textures, resulting in unnatural or distorted outputs~\cite{wulff2020improving}.

\subsection{Artifacts from the Decoding Process}

Even when \(z\) is accurately sampled, decoding artifacts may still arise due to limitations in the network architecture~\cite{karras2020analyzing}. The kernel size and stride used in convolutional layers are particularly influential in determining the fidelity of the output~\cite{karras2021alias}. Large kernels may over-smooth local features, while improper stride configurations can lead to aliasing, both of which degrade image quality.

Moreover, upsampling operations—such as nearest-neighbor or bilinear interpolation—are known to introduce specific artifacts. Nearest-neighbor interpolation often produces jagged edges, whereas bilinear interpolation may blur textures due to its smoothing effect. These operations can significantly impact the realism and perceptual quality of the generated images, especially in high-frequency regions.

\section{Methodology}

\begin{figure*}[t]
    \centering
    \begin{subfigure}{0.16\linewidth}
        \centering
        \includegraphics[width=\linewidth]{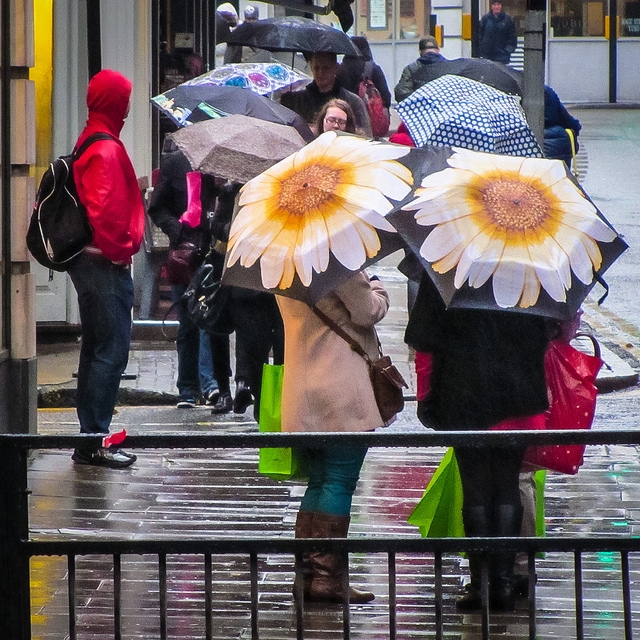}
    \end{subfigure}
    \begin{subfigure}{0.16\linewidth}
        \centering
        \includegraphics[width=\linewidth]{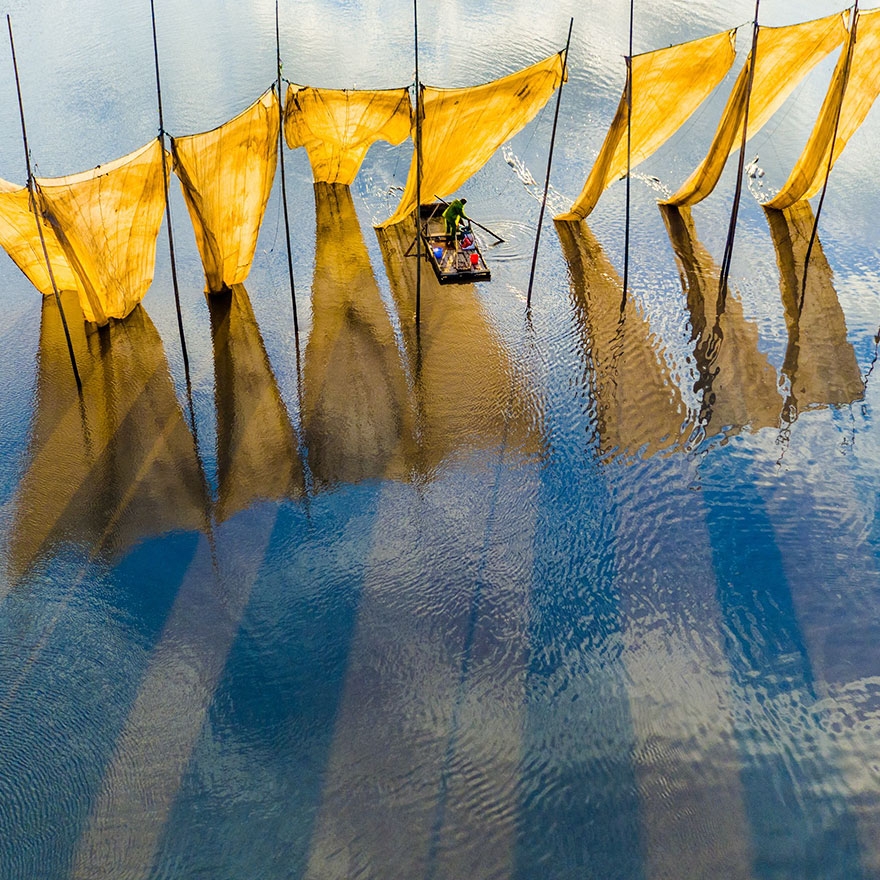}
    \end{subfigure}
    \begin{subfigure}{0.16\linewidth}
        \centering
        \includegraphics[width=\linewidth]{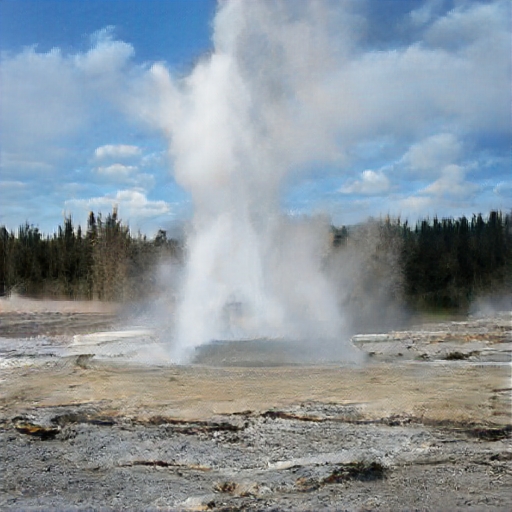}
    \end{subfigure}
    \begin{subfigure}{0.16\linewidth}
        \centering
        \includegraphics[width=\linewidth]{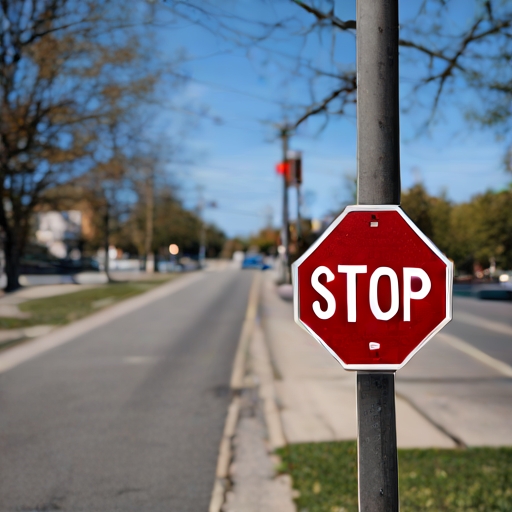}
    \end{subfigure}
     \begin{subfigure}{0.16\linewidth}
        \centering
        \includegraphics[width=\linewidth]{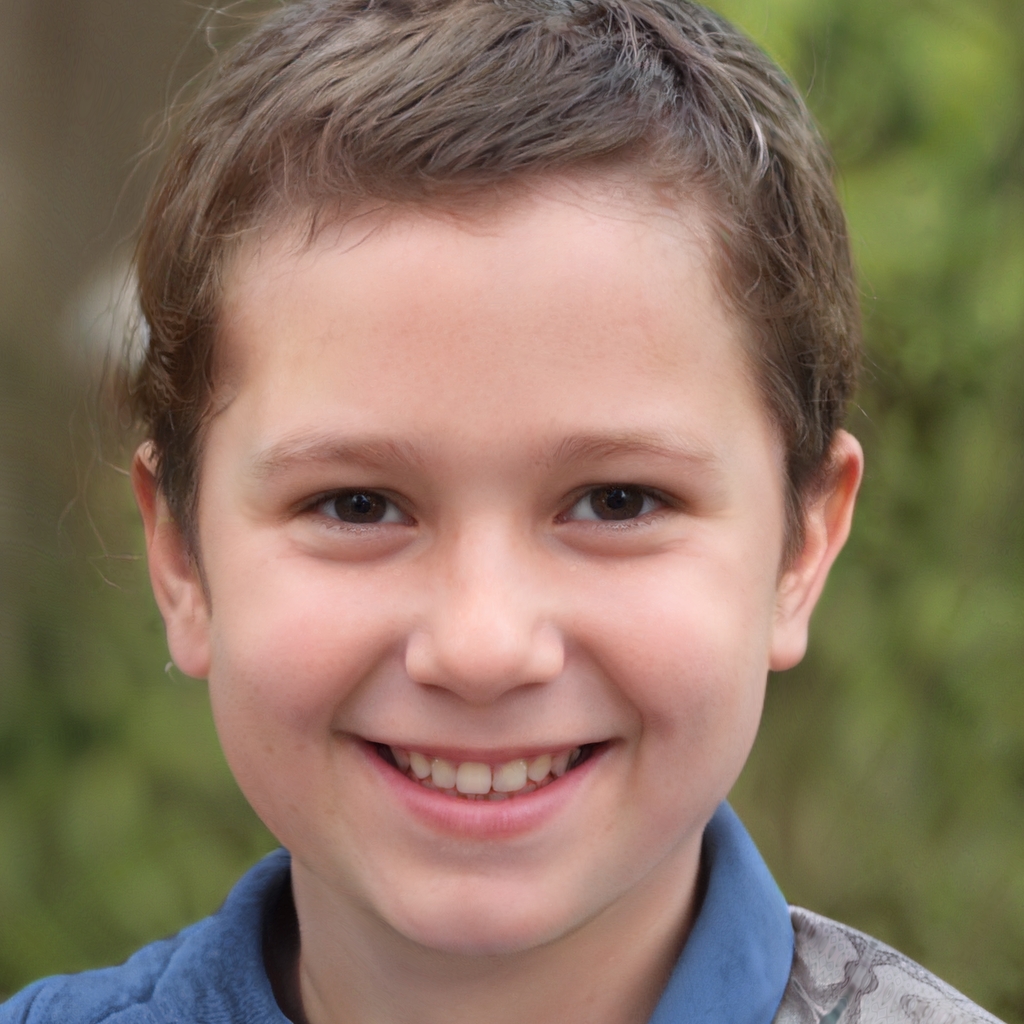}
    \end{subfigure}
    \begin{subfigure}{0.16\linewidth}
        \centering
        \includegraphics[width=\linewidth]{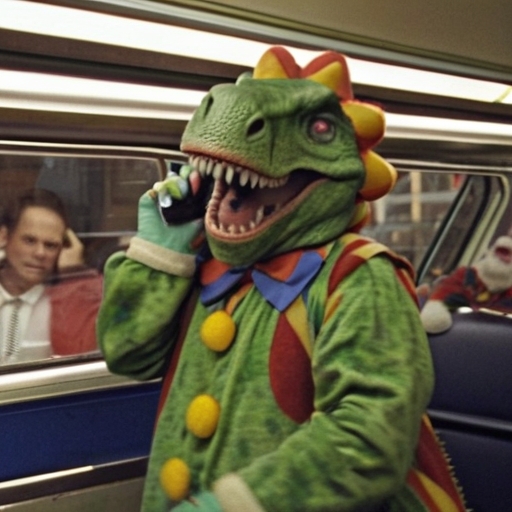}
    \end{subfigure}
    \\
    \begin{subfigure}{0.16\linewidth}
        \centering
        \includegraphics[width=\linewidth]{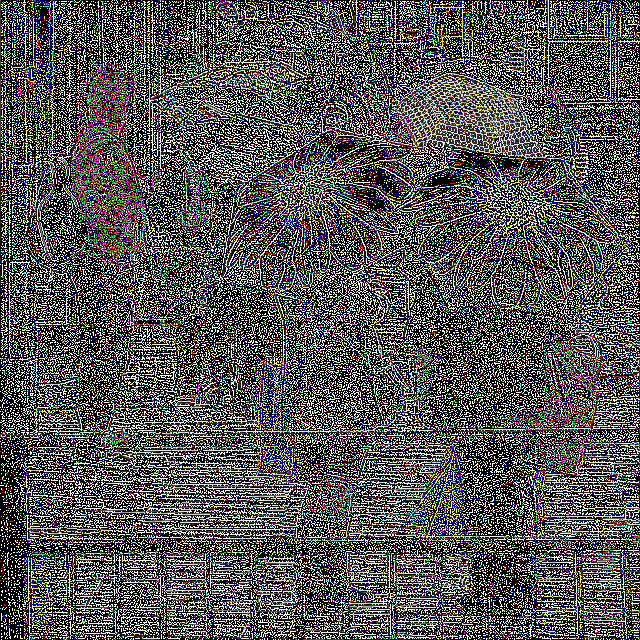}
        \caption{\scriptsize COCO}
    \end{subfigure}
    \begin{subfigure}{0.16\linewidth}
        \centering
        \includegraphics[width=\linewidth]{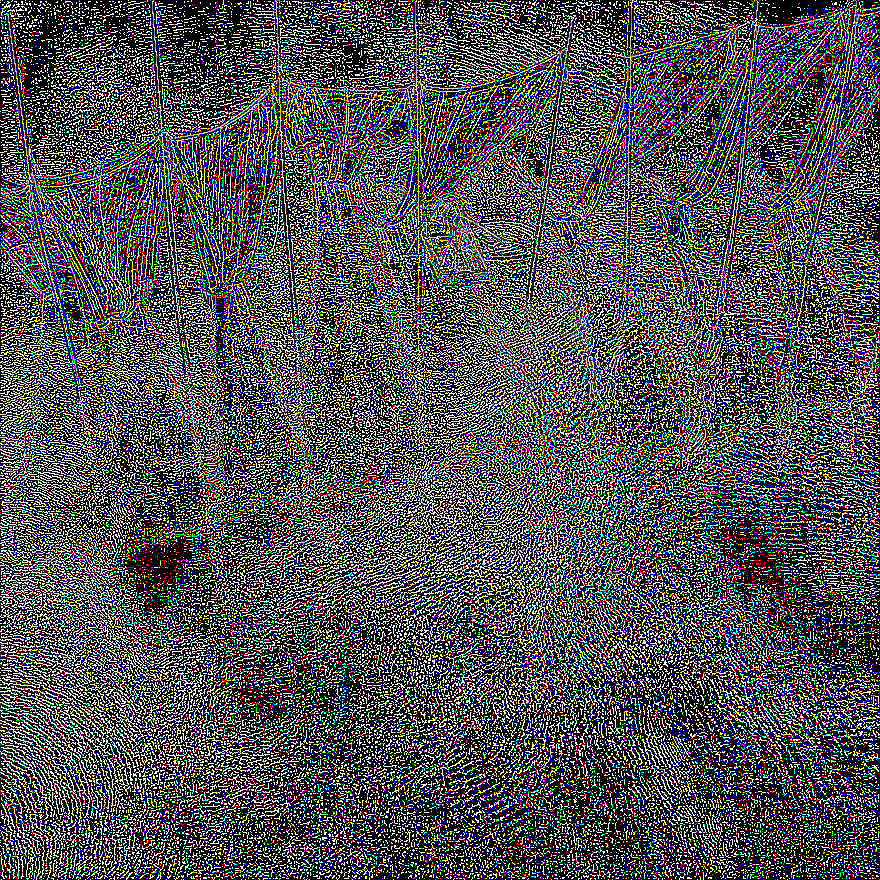}
        \caption{\scriptsize LAION}
    \end{subfigure}
    \begin{subfigure}{0.16\linewidth}
        \centering
        \includegraphics[width=\linewidth]{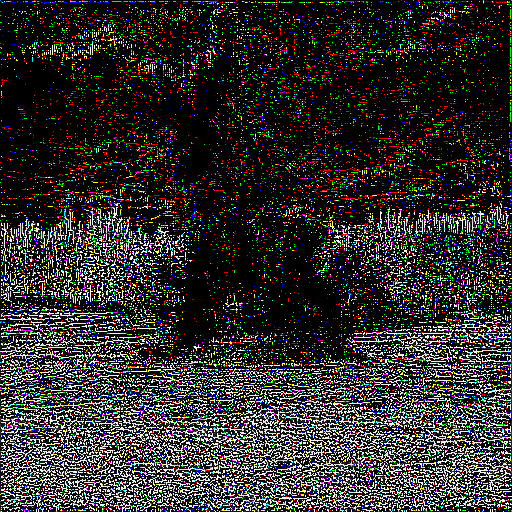}
        \caption{\scriptsize BigGAN}
    \end{subfigure}
    \begin{subfigure}{0.16\linewidth}
        \centering
        \includegraphics[width=\linewidth]{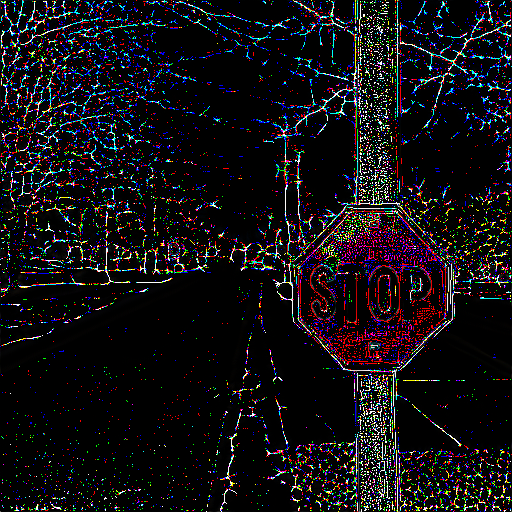}
        {\caption{\scriptsize SDXL-Turbo}}
    \end{subfigure}
    \begin{subfigure}{0.16\linewidth}
        \centering
        \includegraphics[width=\linewidth]{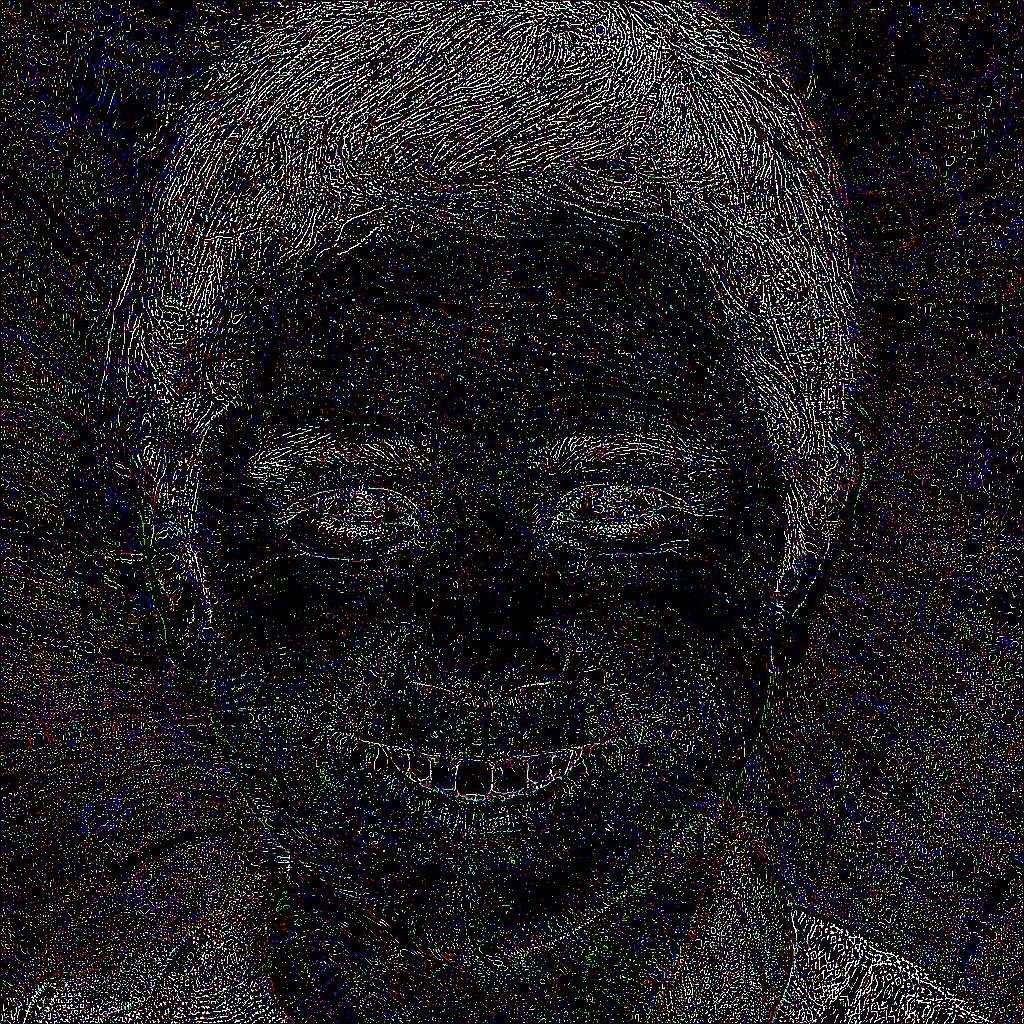}
        \caption{\scriptsize StyleGAN}
    \end{subfigure}
    \begin{subfigure}{0.16\linewidth}
        \centering
        \includegraphics[width=\linewidth]{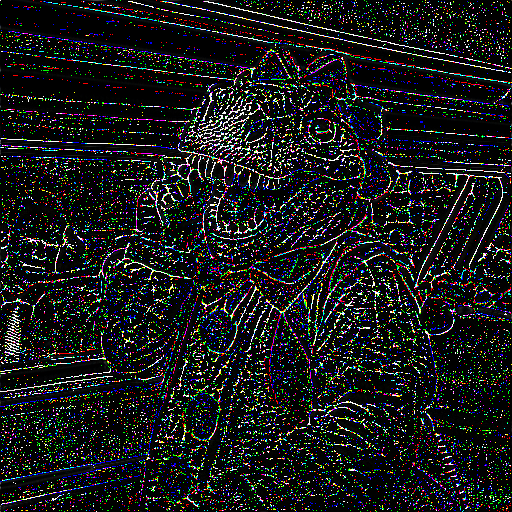}
        \caption{\scriptsize RealVisXL-4.0}
    \end{subfigure}
    \caption{Local pixel dependencies (LPD) comparison between real and synthetic images. Top row: real images (COCO, LAION) and synthetic images (BigGAN, SDXL-Turbo, StyleGAN, RealVisXL-4.0). Bottom row: LPD maps derived from neighborhood-median reconstruction emphasize structural differences.}
    \label{fig:Samples}
\end{figure*}

\subsection{Local Median-based Feature Extraction}
\label{section: Feature}

Natural images exhibit strong local statistical consistency due to the underlying physics of light and matter, where neighboring pixels are highly correlated. Generative models, however, often struggle to perfectly replicate these subtle, complex statistics during the synthesis of high-frequency details. This struggle leads to microscopic disruptions in local pixel dependencies.
We propose a synthetic image detection method based on local statistical dependencies. The core idea is to identify generation artifacts by quantifying the deviation of each pixel from the median of its surrounding neighborhood. The full computational procedure is outlined in Algorithm~\ref{alg:algorithm}.

Let $I$ denote the input image, and $x_{i,j}$ represent the pixel value at location $(i, j)$. According to the Markov Random Field (MRF) assumption, the probability distribution of a pixel depends only on its local neighborhood. Specifically,
\begin{align}
    P(x_{i,j} \mid x_{k,l}, (k,l) \neq (i,j)) 
    = P(x_{i,j} \mid x_{k,l}, (k,l) \in \mathcal{N}_{i,j}),
\end{align}
where $\mathcal{N}_{i,j}$ is the set of neighboring pixels located within an $n \times n$ window centered at $(i,j)$, excluding the center pixel itself:
\begin{align}
    \mathcal{N}_{i,j} = \left\{ x_{k,l} \;\middle|\; 
    \begin{array}{l}
    i - m \leq k \leq i + m,\; j - m \leq l \leq j + m, \\
    (k,l) \ne (i,j)
    \end{array}
    \right\},
\end{align}
with $n = 2m + 1$ and $m \in \mathbb{Z}^{+}$.

To enhance the robustness of the median filtering process and prevent contamination from generated pixels, we introduce a zero-masking strategy that replaces the center pixel with zero before computing the median. This adjustment is particularly beneficial when the neighborhood contains an even number of pixels. The median-based reconstruction at location $(i,j)$ is therefore computed as:
\begin{equation}
    y_{i,j} = \text{Median}(x_{k,l}, (k,l) \in \mathcal{N}'_{i,j}),
\end{equation}
where $\mathcal{N}'_{i,j} = \mathcal{N}_{i,j} \cup \{x_{i,j} = 0\}$ is the extended neighborhood that includes the masked center pixel.

By applying the above operation to all pixels, we obtain a median-reconstructed image $I'$, where each pixel value is replaced by its corresponding $y_{i,j}$. The final local pixel dependency (LPD) feature map is then computed as the pixel-wise difference:
\begin{equation}
    LPD = I - I'.
\end{equation}

Since both $I$ and $I'$ conform to local dependency assumptions, the LPD feature map effectively captures pixel-level inconsistencies and subtle structural deviations, offering strong cues for distinguishing synthetic from natural content, as illustrated in Figure~\ref{fig:Samples}.

\begin{algorithm}[t]
\caption{Local Dependency Feature Extraction via Zero-Masked Median Deviation}
\label{alg:algorithm}
\small
\textbf{Input:} $I$: Image tensor of shape $(C, H, W)$; $n$: Neighborhood size ($n$ is odd)\\
\textbf{Output:} $LPD$: Feature map of shape $(C, H, W)$
\begin{algorithmic}[1]
    \STATE \textcolor{lightgreen}{\# Compute padding size and center index}
    \STATE $p \gets \lfloor n / 2 \rfloor$, \quad $\textit{center\_idx} \gets \lfloor n^2 / 2 \rfloor$
    
    \STATE \textcolor{lightgreen}{\# Pad the image to handle borders}
    \STATE $I_{\text{pad}} \gets \text{Pad}(I, \text{padding}=p, \text{mode}=\text{'constant'}, \text{value}=0)$
    
    \STATE \textcolor{lightgreen}{\# Extract $n \times n$ local patches centered at each pixel}
    \STATE $I_{\text{patches}} \gets \text{Unfold}(I_{\text{pad}}, \text{kernel\_size}=n)$ \quad \textcolor{lightgreen}{// shape: $(C, n^2, H \cdot W)$}
    
    \STATE \textcolor{lightgreen}{\# Zero out the center pixel in each patch}
    \STATE $I_{\text{patches}}[:, \textit{center\_idx}, :] \gets 0$
    
    \STATE \textcolor{lightgreen}{\# Compute median along patch dimension}
    \STATE $I'_{\text{med}} \gets \text{Median}(I_{\text{patches}}, \text{dim}=1)$
    
    \STATE \textcolor{lightgreen}{\# Reshape to match original image dimensions}
    \STATE $I' \gets \text{Reshape}(I'_{\text{med}}, \text{shape}=(C, H, W))$
    
    \STATE \textcolor{lightgreen}{\# Compute local pixel dependency map}
    \STATE $LPD \gets I - I'$
    
    \STATE \textbf{return} $LPD$
\end{algorithmic}
\end{algorithm}

This method effectively integrates the local dependency modeling capabilities of Markov Random Fields with the robustness of median filtering, providing a principled and resilient strategy for detecting subtle inconsistencies in synthetic imagery.

\subsection{FerretNet Architecture}
\begin{figure*}[t]
\centering
\includegraphics[width=1.0\textwidth]{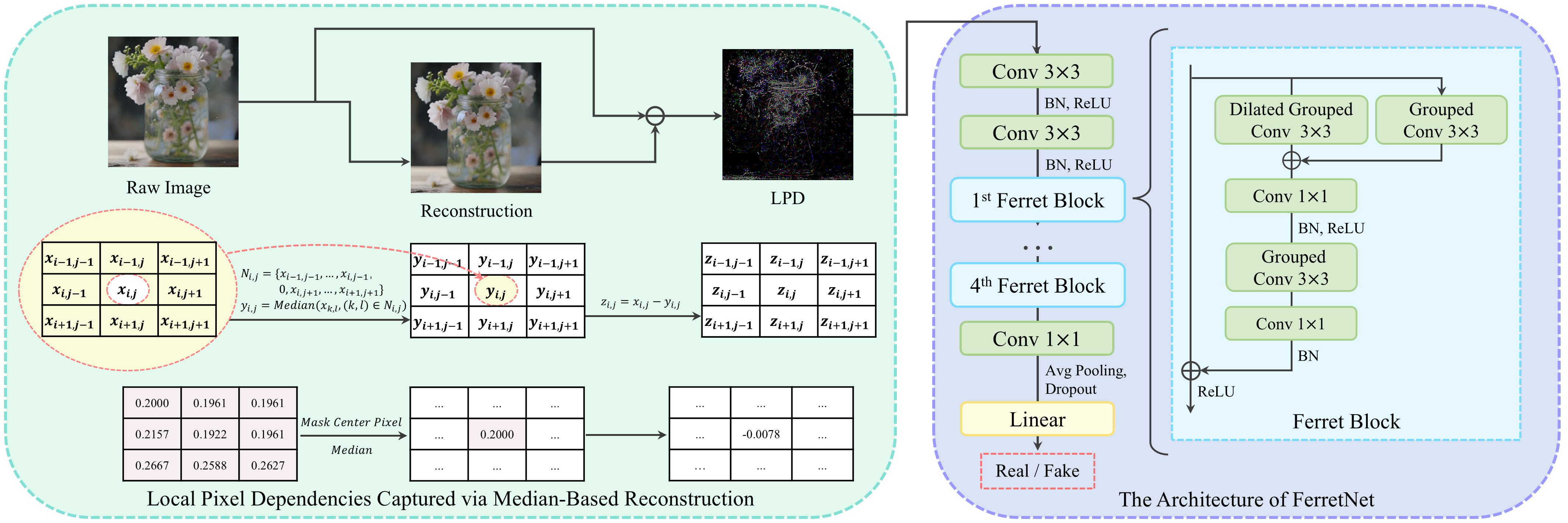}
\caption{Pipeline of FerretNet: computation of local pixel median discrepancy for artifact representation, followed by lightweight detection using depthwise separable and dilated convolutions.}
\label{fig:FerretNet}
\end{figure*}

FerretNet is a lightweight convolutional neural network designed to achieve a balance between computational efficiency and feature extraction capability. As illustrated in Figure~\ref{fig:FerretNet}, the network begins with two conventional $3\times3$ convolutional layers for initial feature extraction, each followed by Batch Normalization (BN)~\cite{ioffe2015batch} and ReLU activation. 

At the core of FerretNet are four cascaded Ferret Blocks, which progressively refine the extracted features while keeping the model compact. The final stage comprises a $1\times1$ convolution, global average pooling, Dropout regularization, and a fully connected layer for classification.

The key innovation lies in the Ferret Block, which is designed to expand the effective receptive field under constrained network depth, thereby enhancing the model’s capacity for local pattern extraction. Each Ferret Block adopts a dual-path parallel architecture to increase the receptive field:

\begin{itemize}
    \item The \textbf{primary path} employs a $3\times3$ \emph{dilated grouped convolution} with a dilation rate of 2. The number of groups equals the number of input channels, allowing the receptive field to expand without increasing the number of parameters.
    \item The \textbf{secondary path} utilizes a standard $3\times3$ \emph{grouped convolution}, maintaining the same grouping structure to capture fine-grained local patterns.
\end{itemize}

This dual-path configuration approximates a sparse $5\times5$ receptive field via parallel processing, enabling FerretNet to simulate deeper network behaviors within shallower layers, thus reducing computational cost. The outputs from both paths are fused through a $1\times1$ convolution, followed by BN and ReLU activation. Additional $3\times3$ grouped and $1\times1$ convolution layers further enrich the feature representation. Residual connections are employed to facilitate stable gradient propagation and enhance learning stability.

\section{Experiments}

\subsection{Dataset Construction}

\subsubsection{Training Dataset}
To ensure a consistent evaluation baseline, we follow the protocols established in~\cite{jeong2022bihpf,ojha2023towards,liu2024forgery,tan2024rethinking}, utilizing four semantic classes (car, cat, chair, horse) from the ForenSynths dataset~\cite{wang2020cnn}. Each class contains 18,000 synthetic images generated by ProGAN~\cite{karras2018progressive}, paired with an equal number of real images from the LSUN dataset~\cite{yu2015lsun}. All methods compared in this study were trained or fine-tuned on this same limited ProGAN 4-class dataset, except for CO-SPY~\cite{cheng2025co}, which utilized its officially released weights trained on other datasets.

\subsubsection{Testing Dataset}

To assess the generalization ability of the proposed method under real-world conditions, we evaluate its performance on diverse synthetic and real images from four distinct test sets, comprising a total of 22 generative models:

\textbf{ForenSynths.} This test set includes synthetic images generated by eight representative generative models: ProGAN~\cite{karras2018progressive}, StyleGAN~\cite{Karras_2019_CVPR}, StyleGAN2~\cite{karras2020analyzing}, BigGAN~\cite{brock2018large}, CycleGAN~\cite{zhu2017unpaired}, StarGAN~\cite{choi2018stargan}, GauGAN~\cite{park2019gaugan}, and Deepfake~\cite{rossler2019faceforensics}. Real images are sourced from six widely-used datasets: LSUN~\cite{yu2015lsun}, ImageNet~\cite{russakovsky2015imagenet}, CelebA~\cite{liu2015celeba}, CelebA-HQ~\cite{karras2018celebahq}, COCO~\cite{lin2014microsoft}, and FaceForensics++~\cite{rossler2019faceforensics}, totaling 62,000 images.

\textbf{Diffusion-6-cls.} As described in FatFormer~\cite{liu2024forgery}, this test set comprises synthetic images generated by six diffusion-based models collected from DIRE~\cite{wang2023dire} and Ojha et al.~\cite{ojha2023towards}, including DALL-E~\cite{ramesh2021dalle}, Guided~\cite{dhariwal2021guide}, PNDM~\cite{liu2022pndm}, VQ-Diffusion~\cite{gu2022vqdiffusion}, Glide~\cite{nichol2022glide}, and LDM~\cite{rombach2022ldm}. Variants produced by Glide and LDM with different parameter configurations are treated as separate categories (see original papers for details). Each subset includes 1,000 synthetic and 1,000 real images, with some real images reused across subsets.

\textbf{Synthetic-Pop.} To capture the latest progress in high-resolution image generation, we constructed the Synthetic-Pop dataset using six popular models—Openjourney~\cite{Openjourney}, Proteus-0.3~\cite{proteus_v03}, RealVisXL-4.0~\cite{RealVisXL_V4}, SD-3.5-Medium~\cite{esser2024sd35}, SDXL-Turbo~\cite{podell2024sdxl}, and YiffyMix~\cite{YiffyMix}. Each model was prompted with 5,000 captions randomly sampled from COCO~\cite{lin2014microsoft}. Real images were drawn from COCO and LAION-Aesthetics V2 (6.5+)~\cite{LaionAesthetics}, resulting in six subsets, each containing 5,000 synthetic and 5,000 real images (60,000 images total).

\textbf{Synthetic-Aesthetic.} To further investigate the aesthetic and stylistic diversity of synthetic imagery, we sampled 40,000 images from the Simulacra Aesthetic Captions (SAC) dataset~\cite{pressmancrowson2022}, which were generated by CompVis latent GLIDE~\cite{rombach2022ldm} and Stable Diffusion~\cite{rombach2022ldm} using prompts sourced from over 40,000 real users. An equal number of real images were sampled from LAION-Aesthetics V2 (6.5+)~\cite{LaionAesthetics}, resulting in a total of 80,000 images. This dataset provides a challenging benchmark for evaluating performance under realistic and user-driven conditions.

\subsection{Implementation Details}

FerretNet is trained from scratch without any pretraining. We use the Adam optimizer with a learning rate of $2\times10^{-4}$, betas of $(0.937, 0.999)$, and a weight decay of $5\times10^{-4}$. The model is trained for 100 epochs using a batch size of 32.
During training, input images are randomly cropped to a resolution of $224\times224$ and augmented with random horizontal flipping. Binary Cross Entropy with Logits Loss (BCEWithLogitsLoss) is adopted as the loss function. For evaluation, images are center-cropped to $256\times256$.

Following previous work~\cite{jeong2022bihpf,ojha2023towards,liu2024forgery}, Accuracy (ACC) and Average Precision (AP) are used as the primary evaluation metrics. To measure real-world performance, we report throughput on the Synthetic-Aesthetic test set using an NVIDIA RTX 4090 GPU and an Intel(R) Xeon(R) Gold 6430 CPU (16 vCPUs), with a batch size of 128.

\subsection{Main Results}

\begin{table}[htbp]
\caption{Accuracy and average precision comparisons with peer methods on ForenSynth test set for different GAN images and Deepfake images. The best and second best performance are highlighted in \textbf{bold} and \underline{underlined}, respectively.}
\scriptsize
\centering
\begin{tabular}{
p{1.6cm}|
>{\centering\arraybackslash}p{0.9cm}
>{\centering\arraybackslash}p{0.9cm}
>{\centering\arraybackslash}p{0.9cm}
>{\centering\arraybackslash}p{0.9cm}
>{\centering\arraybackslash}p{0.9cm}
>{\centering\arraybackslash}p{0.9cm}
>{\centering\arraybackslash}p{0.9cm}
>{\centering\arraybackslash}p{0.9cm}|
>{\centering\arraybackslash}p{0.8cm}
}
\toprule 
Methods & ProGAN & StyleGAN & StyleGAN2 & BigGAN & CycleGAN & StarGAN & GauGAN & Deepfake & Mean \\ 
\midrule
Wang~\cite{wang2020cnn} & 91.4/99.4 & 63.8/91.4 & 76.4/97.5 & 52.9/73.3 & 72.7/88.6 & 63.8/90.8 & 63.9/92.2 & 51.7/62.3 & 67.1/86.9\\
F3Net~\cite{qian2020thinking} & 99.4/100.0 & 92.6/99.7 & 88.0/99.8 & 65.3/69.9 & 76.4/84.3 & 100.0/100.0 & 58.1/56.7 & 63.5/78.8 & 80.4/86.2\\
FrePGAN~\cite{jeong2022frepgan} & 99.0/99.9 & 80.7/89.6 & 84.1/98.6 & 69.2/71.1 & 71.1/74.4 & 99.9/100.0 & 60.3/71.7 & 70.9/91.9 & 79.4/87.2\\
BiHPF~\cite{jeong2022bihpf} & 90.7/86.2 & 76.9/75.1 & 76.2/74.7 & 84.9/81.7 & 81.9/78.9 & 94.4/94.4 & 69.5/78.1 & 54.4/54.6 & 78.6/77.9\\
LGrad~\cite{tan2023learning} & 99.9/100.0 & 94.8/99.9 & 96.0/99.9 & 82.9/90.7 & 85.3/94.0 & 99.6/100.0 & 72.4/79.3 & 58.0/67.9 & 86.1/91.5\\
Ojha~\cite{ojha2023towards} & 99.7/100.0 & 89.0/98.7 & 83.9/98.4 & 90.5/99.1 & 87.9/99.8 & 91.4/100.0 & 89.9/100.0 & 80.2/90.2 & 89.1/98.3\\
FreqNet~\cite{tan2024frequency} & 99.6/100.0 & 90.2/99.7 & 88.0/99.5 & 90.5/96.0 & 95.8/99.6 & 85.7/99.8 & 93.4/98.6 & 88.9/94.4 & 91.5/98.5\\
NPR~\cite{tan2024rethinking} & 99.8/100.0 & 96.3/99.8 & 97.3/100.0 & 87.5/94.5 & 95.0/99.5 & 99.7/100 & 86.6/88.8 & 77.4/86.2 & 92.5/96.1\\
FatFormer~\cite{liu2024forgery} & 99.9/100.0 & 97.2/99.8 & 98.8/99.9 & 99.5/100.0 & 99.3/100.0 & 99.8/100.0 & 99.4/100.0 & 93.2/98.0 & \textbf{98.4/99.7}\\
SAFE~\cite{li2025improving} & 99.9/100.0 & 98.0/99.9 & 98.6/100.0 & 89.7/95.9 & 98.9/99.8 & 99.9/100.0 & 91.5/97.2 & 93.1/97.5 & \underline{96.2/98.8} \\
CO-SPY~\cite{cheng2025co} & 74.7/78.1 & 63.9/70.2 & 59.7/63.1 & 71.6/83.9 & 58.5/55.8 & 62.1/94.3 & 69.6/83.4 & 65.7/79.7 & 65.7/76.1 \\
\midrule
FerretNet (Our) & 99.9/100.0 & 98.0/100.0 & 98.5/100.0 & 92.6/98.5 & 98.8/99.9 & 99.1/100.0 & 91.4/99.8 & 89.2/96.7 & 95.9/99.3 \\   
\bottomrule
\end{tabular}
\label{table: ForenSynth Results}
\end{table}

\begin{table}[htbp]
\caption{Accuracy and average precision comparisons with peer methods on Diffusion-6-cls test set.}
\centering
\scriptsize
\begin{tabular}{
p{1.6cm}|
>{\centering\arraybackslash}p{0.8cm}
>{\centering\arraybackslash}p{0.8cm}
>{\centering\arraybackslash}p{0.8cm}
>{\centering\arraybackslash}p{0.9cm}
>{\centering\arraybackslash}p{0.8cm}
>{\centering\arraybackslash}p{1.2cm}
>{\centering\arraybackslash}p{0.8cm}
>{\centering\arraybackslash}p{1.2cm}|
>{\centering\arraybackslash}p{0.9cm}
}
\toprule
Dataset & Wang~\cite{wang2020cnn} & LGrad~\cite{tan2023learning} & Ojha~\cite{ojha2023towards} & FreqNet~\cite{tan2024frequency} & NPR~\cite{tan2024rethinking} & FatFormer~\cite{liu2024forgery} & SAFE~\cite{li2025improving} & CO-SPY~\cite{cheng2025co} & FerretNet \\
\midrule
Dall-E             & 51.8/61.3 & 88.5/97.3  & 89.5/96.8 & 97.3/99.7   & 90.9/98.1   & 98.8/99.8   & 97.5/99.7   & 81.8/87.2 & 91.4/98.2 \\
Guided             & 54.9/66.6 & 86.6/100.0 & 75.7/85.1 & 67.2/75.4   & 74.0/78.1   & 76.1/92.0   & 82.4/95.8   & 62.5/86.0 & 92.1/98.6 \\
PNDM               & 50.8/90.3 & 69.8/98.5  & 75.3/92.5 & 85.2/99.9   & 97.5/100.0  & 99.3/100.0  & 78.9/98.6   & 53.0/55.6 & 96.9/100.0 \\
VQ-Diffusion       & 50.0/71.0 & 96.3/100.0 & 83.5/97.7 & 100.0/100.0 & 100.0/100.0 & 100.0/100.0 & 100.0/100.0 & 71.9/71.5 & 99.9/100.0 \\
Glide-50-27        & 54.2/76.0 & 90.7/95.1  & 91.1/97.4 & 86.6/95.8   & 97.5/99.5   & 94.7/99.4   & 96.6/99.2   & 69.1/74.6 & 97.2/99.7 \\
Glide-100-10       & 53.3/72.9 & 89.4/94.9  & 90.1/97.0 & 87.8/96.0   & 97.8/99.5   & 94.2/99.2   & 97.3/99.4   & 76.6/81.6 & 97.9/99.9 \\
Glide-100-27       & 53.0/71.3 & 87.4/93.2  & 90.7/97.2 & 84.4/95.6   & 97.4/99.5   & 94.4/99.1   & 95.8/98.9   & 73.5/78.2 & 97.3/99.7 \\
LDM-100            & 51.9/63.7 & 94.8/99.2  & 90.5/97.0 & 97.8/99.9   & 98.0/99.6   & 98.7/99.9   & 98.8/100.0  & 82.7/86.9 & 98.8/100.0 \\
LDM-200            & 52.0/64.5 & 94.2/99.1  & 90.2/97.1 & 97.4/99.9   & 98.2/99.6   & 98.6/99.8   & 98.8/100.0  & 83.1/87.5 & 98.8/100.0 \\
LDM-200-CFG        & 51.6/63.1 & 95.9/99.2  & 77.3/88.6 & 97.3/99.9   & 98.0/99.5   & 94.9/99.1   & 98.7/99.9   & 85.3/91.0 & 98.5/99.9 \\
\midrule
Mean               & 52.4/70.1 & 89.4/97.7  & 85.4/94.6 & 90.1/96.2 & 94.9/97.3 & \underline{95.0/98.8} & 94.5/99.1 & 73.9/80.0 & \textbf{96.9/99.6} \\
\bottomrule
\end{tabular}
\label{table: diffusion-6-cls results}
\end{table}

\begin{table}[htbp]
\caption{Accuracy and average precision comparisons with state-of-the-art methods on Synthetic-Pop test set.}
\scriptsize
\centering
\begin{tabular}{
p{1.4cm}|
>{\centering\arraybackslash}p{1.2cm}
>{\centering\arraybackslash}p{1.2cm}
>{\centering\arraybackslash}p{1.5cm}
>{\centering\arraybackslash}p{1.7cm}
>{\centering\arraybackslash}p{1.4cm}
>{\centering\arraybackslash}p{1.1cm}|
>{\centering\arraybackslash}p{1.1cm}
}
\toprule
Methods & Openjourney & Proteus-0.3 & RealVisXL-4.0 & SD-3.5-Medium & SDXL-Turbo & YiffyMix & Mean \\
\midrule
FreqNet~\cite{tan2024frequency} & 56.3 / 63.6 & 44.0 / 41.2 & 59.4 / 66.6 & 78.5 / 86.8 & 77.5 / 86.0 & 74.3 / 84.4 & 65.0 / 71.4 \\
NPR~\cite{tan2024rethinking} & 78.8 / 83.5 & 68.6 / 69.3 & 78.1 / 82.0 & 80.4 / 84.1 & 78.2 / 82.9 & 80.0 / 85.1 & 77.4 / 81.2 \\
FatFormer~\cite{liu2024forgery} & 58.8 / 65.4 & 93.9 / 97.6 & 49.0 / 41.7  & 81.9 / 89.1 & 58.7 / 65.3 & 80.9 / 89.9 & 70.5 / 74.8 \\
SAFE~\cite{li2025improving} & 94.7 / 99.3 & 99.2 / 99.9 & 97.9 / 99.8 & 98.1 / 99.7 & 98.1 / 99.8 & 99.5 / 99.9 & \underline{97.9 / 99.7} \\
CO-SPY~\cite{cheng2025co} & 92.4 / 97.6 & 88.8 / 93.0 & 79.0 / 86.5 & 80.9 / 87.8 & 79.9 / 88.3 & 92.9 / 97.5 & 85.6 / 91.8 \\
\midrule
FerretNet & 98.4 / 99.7 & 98.6 / 99.7 & 98.8 / 99.9 & 97.2 / 99.6 & 98.9 / 100.0 & 97.8 / 99.7 & \textbf{98.3 / 99.8} \\
\bottomrule
\end{tabular}
\label{table: Synthetic-Pop results}
\end{table}

\begin{table}[htbp]
\caption{Performance comparisons with state-of-the-art methods across four distinct test sets. Throughput measurements were conducted on the Synthetic-Aesthetic test set. Upward arrows indicate that higher values are better, while downward arrows signify the opposite.}
\scriptsize
\centering
\begin{tabular}{
p{1.8cm}|
>{\centering\arraybackslash}p{1.5cm}
>{\centering\arraybackslash}p{1.5cm}
>{\centering\arraybackslash}p{1.5cm}
>{\centering\arraybackslash}p{1.5cm}
>{\centering\arraybackslash}p{1.5cm}
>{\centering\arraybackslash}p{1.5cm}
}
\toprule
Methods & Ref & Image size & Params (M) $\downarrow$ & FLOPs (G) $\downarrow$ & FPS $\uparrow$ & ACC / AP $\uparrow$ \\
\midrule
FreqNet~\cite{tan2024frequency} & AAAI 2024 & 256$^2$ & 1.85 & 2.58 & 200.2 & 79.2 / 86.8 \\
NPR~\cite{tan2024rethinking} & CVPR 2024 & 256$^2$ & \underline{1.44} & \textbf{2.29} & \underline{720.9} & 86.5 / 89.4 \\
FatFormer~\cite{liu2024forgery} & CVPR 2024 & 224$^2$ & 492.59 & 269.92 & 88.6 & 86.1 / 91.0 \\
SAFE~\cite{li2025improving} & KDD 2025 & 256$^2$ & \underline{1.44} & \textbf{2.29} & 770.2 & \underline{96.8 / 99.3} \\
CO-SPY~\cite{cheng2025co} & CVPR 2025 & 384$^2$ & 963.05 & 644.80 & 26.3 & 76.5 / 83.8 \\
\midrule
\textbf{FerretNet (Ours)} & - & 256$^2$ & \textbf{1.06} & \underline{2.38} & \textbf{772.1} & \textbf{97.1 / 99.6} \\
\bottomrule
\end{tabular}
\label{table: Summary}
\end{table}

We begin by evaluating FerretNet on GAN-based and Deepfake images using the ForenSynths test set. As shown in Table~\ref{table: ForenSynth Results}, it achieves an average accuracy (ACC) of 95.9\%, outperforming lightweight baselines such as FreqNet~\cite{tan2024frequency} (91.5\%) and NPR~\cite{tan2024rethinking} (92.5\%). Although FatFormer~\cite{liu2024forgery} reports a higher ACC of 98.4\%, it relies on pre-trained CLIP weights, whereas FerretNet achieves competitive accuracy with significantly fewer parameters.

Next, on diffusion-generated images (Table~\ref{table: diffusion-6-cls results}), FerretNet attains an ACC of 96.9\% and an AP of 99.6\%, outperforming FatFormer~\cite{liu2024forgery} by 1.9 and 0.8 percentage points (pp), respectively. Other lightweight models such as NPR~\cite{tan2024rethinking}, FreqNet~\cite{tan2024frequency} and SAFE~\cite{li2025improving} perform less favorably, with ACC scores falling below 95.0\%.

We further evaluate performance on high-quality synthetic images using the Synthetic-Pop test set (Table~\ref{table: Synthetic-Pop results}). Some existing methods experience noticeable degradation; for example, NPR~\cite{tan2024rethinking} achieves only 77.4\% ACC and 81.2\% AP. In contrast, FerretNet maintains 98.3\% ACC and 99.8\% AP, highlighting its robustness and reliability on visually realistic forgeries.

To evaluate real-world applicability, we tested FerretNet on four distinct test sets for both detection performance and efficiency. As shown in Table~\ref{table: Summary}, FerretNet achieves 97.1\% ACC and 99.6\% AP with 1.06M parameters and 772.1 FPS on an RTX 4090. Notably, it outperforms FatFormer~\cite{liu2024forgery} by 11.0 and 8.6 pp in ACC and AP, respectively, while using only 0.2\% of its parameters.

Finally, Appendix~\ref{sec: visualization analysis} provides a detailed analysis of specific success and failure cases, further clarifying the model's decision boundaries.

\subsection{Ablation Study}

Unless specified, all ablation results report the average ACC and AP across four datasets: ForenSynths, Diffusion-6-cls, Synthetic-Pop, and Synthetic-Aesthetic. More supplementary experiments see Appendix~\ref{sec: add exp}.

\subsubsection{Impact of Different Local Neighborhood Sizes}

\begin{table*}[htbp]
\caption{Impact of the local neighborhood size.}
\scriptsize
\centering
\begin{tabular}{l|ccc|cccc|c}
\toprule
\multirow{2}{*}{\makecell[b]{Inputs}} & \multicolumn{3}{c|}{\makecell[c]{Size $(n\times n)$}} & \multicolumn{5}{c}{\makecell[c]{ACC / AP on the Test set}} \\
\cmidrule(lr){2-4} \cmidrule(lr){5-9}
& $3\times3$ & $5\times5$ & $7\times7$ & ForenSynths & Diffusion-6-cls & Synthetic-Pop & Synthetic-Aesthetic & Mean\\
\midrule
$I$          & & &            & 84.6 / 88.9 & 87.8 / 96.8 & 84.5 / 92.9  & 90.5 / 95.3 & 86.9 / 93.5 \\
$LPD$ & \checkmark & & & \textbf{95.9 / 99.3} & \textbf{96.9 / 99.6} & \textbf{98.3 / 99.8} & \textbf{97.3 / 99.6} & \textbf{97.1 / 99.6} \\
$LPD$ & & \checkmark & & \underline{91.8 / 96.2} & \underline{95.8 / 99.3} & \underline{91.1 / 97.4} & \underline{96.9 / 98.9} & \underline{93.9 / 98.0} \\
$LPD$ & & & \checkmark & 82.4 / 90.6 & 85.2 / 93.6 & 78.6 / 91.9 & 85.0 / 94.4 & 82.8 / 92.6 \\
\bottomrule
\end{tabular}
\label{table: neighborhood size}
\end{table*}

Table~\ref{table: neighborhood size} shows that $LPD$ extracted using a $3 \times 3$ local neighborhood substantially enhances detection accuracy compared to raw input $I$. Average ACC improves from 86.9\% to 97.1\% (+10.2\%), and AP rises from 93.5\% to 99.6\% (+6.1\%). However, performance deteriorates as the neighborhood size increases. For instance, using a $7 \times 7$ neighborhood weakens feature discrimination and significantly reduces detection accuracy.

This trend aligns with the structural characteristics of generative models, which typically employ $2\times$ upsampling and small convolutional kernels ($1 \times 1$ or $3 \times 3$). The $3 \times 3$ neighborhood is particularly effective in capturing localized decoding artifacts for two reasons:  
1) It matches the scale of operations used in generative architectures, making it ideal for exposing subtle synthesis artifacts;  
2) It captures local pixel variations while suppressing potential noise artifacts.

\subsubsection{Impact of Center Pixel Processing Methods}

According to Section~\ref{section: Feature}, the neighborhood median \(y_{i,j}\) should satisfy two key requirements: reducing the interference of center pixels in median computation, and ensuring the median value equals a real pixel value from the neighborhood set when possible, thus enhancing statistical correlation with the original image. To validate the effectiveness of zero-value masking strategy, we compared three center pixel processing methods:

\begin{wraptable}{r}{0.5\textwidth}
\vspace{-1.35em}
\caption{Impact of center pixel processing methods on metrics (ACC/AP): average results across varying local neighborhood sizes.}
\centering
\scriptsize
\begin{tabular}{l|ccc}
\toprule 
Methods & $3\times 3$ & $5\times 5$ & $7\times 7$ \\  
\midrule
Mask      & \textbf{97.1 / 99.6}      & \textbf{93.9 / 98.0}    & 82.8 / 92.6             \\
Exclusion & \underline{95.3 / 98.8}   & \underline{90.5 / 96.3} & \underline{86.4 / 93.6} \\
Retention & 93.3 / 97.6               & 89.7 / 96.3             & \textbf{87.5 / 93.1}    \\
\bottomrule
\end{tabular}
\label{table: center pixel}
\vspace{-1.0em}
\end{wraptable}

1. \textbf{Zero-value Masking}: Set the center pixel to zero while keeping it in the set. This increases the probability that the median equals a real neighborhood pixel and reduces the center pixel's influence.

2. \textbf{Complete Exclusion}: Remove the center pixel entirely. This results in a non-existent pixel value (i.e., not from the original image), thereby weakening the dependency on the source image.

3. \textbf{Center Pixel Retention}: Keep the original center pixel, as in standard median filtering. This approach compromises the ability to detect local anomalies.

The experimental results in Table~\ref{table: center pixel} demonstrate that for local neighborhood sizes of $3\times3$ and $5\times5$, the zero-value masking achieves the highest detection accuracy, followed by the complete exclusion, with the center pixel retention yielding the lowest accuracy. These findings validate the effectiveness of the proposed strategy.

\subsubsection{Impact of Neighborhood Statistic Selection}

\begin{wraptable}{r}{0.5\textwidth}
\vspace{-1.35em}
\caption{Impact of neighborhood statistic selection methods.}
\centering
\scriptsize
\begin{tabular}{l|ccc}
\toprule
Methods & $3\times 3$ & $5\times 5$ & $7\times 7$ \\  
\midrule
Max & \underline{93.6 / 97.9} & 86.8 / 94.3             & \underline{88.9 / 94.8} \\
Avg & 92.2 / 97.2             & 88.2 / 94.4             & \textbf{90.0 / 96.6}    \\
Min & 91.8 / 96.9             & \underline{88.3 / 94.7} & 87.6 / 94.0             \\
Med & \textbf{97.1 / 99.6}    & \textbf{93.9 / 98.0}    & 82.8 / 92.6             \\
\bottomrule
\end{tabular}
\label{table: pixel selection}
\end{wraptable}

To verify the advantages of the neighborhood median-based feature extraction strategy in synthetic image detection, we designed three alternative methods: selecting the maximum, minimum, and average values from the neighborhood. The center pixel was masked by setting it to infinity, negative infinity, or zero, respectively, to reduce its influence on feature extraction.
The experimental results in Table~\ref{table: pixel selection} show that, for both $3 \times 3$ and $5 \times 5$ local neighborhoods, the median strategy significantly outperforms the other methods.

\subsubsection{Impact of Different Backbones}

\begin{wraptable}{r}{0.5\textwidth}
\vspace{-2.35em}
\caption{Comparison of different backbones with and without $LPD$ as input.}
\centering
\scriptsize
\begin{tabular}{l|c|c|c|c}
\toprule
Methods                   & Params                  & w / o      & Throughput $\uparrow$ & ACC / AP $\uparrow$ \\
\midrule
\multirow{2}{*}{Xception} & \multirow{2}{*}{20.8 M} & ×          & 730.5 Img/s           & 89.8 / 94.1          \\
                          &                         & \checkmark & 710.6 Img/s           & \textbf{95.1 / 98.8} \\ 
\midrule
\multirow{2}{*}{ResNet50} & \multirow{2}{*}{23.5 M} & ×          & 755.4 Img/s           & 75.0 / 80.3          \\
                          &                         & \checkmark & 750.9 Img/s           & \textbf{81.1 / 85.6} \\
\midrule
\multirow{2}{*}{FerretNet} & \multirow{2}{*}{1.1 M} & ×          & 777.8 Img/s           & 86.9 / 93.5          \\
                           &                        & \checkmark & 772.1 Img/s           & \textbf{97.1 / 99.6} \\
\bottomrule
\end{tabular}
\label{table: Backbone results}
\end{wraptable}

We evaluated ResNet50~\cite{he2016deep}, Xception~\cite{chollet2017xception}, and our proposed FerretNet on both raw image $I$ and $LPD$ inputs.
As shown in Table~\ref{table: Backbone results}, FerretNet achieves competitive accuracy on raw images despite having significantly fewer parameters, and outperforms the other architectures when leveraging $LPD$.
Across all backbones, replacing $I$ with $LPD$ consistently delivers accuracy gains with negligible effect on inference speed.

\section{Conclusion}
This work presents a universal artifact representation framework and introduces FerretNet, a lightweight yet effective neural network for synthetic image detection. FerretNet achieves a remarkable 99.8\% reduction in parameters compared to the state-of-the-art method FatFormer~\cite{liu2024forgery}, while maintaining exceptional detection accuracy, reaching 97.1\% on images generated by 22 different generative models. It demonstrates strong generalization capabilities and computational efficiency, outperforming existing approaches on high-quality synthetic datasets. Our contributions include a novel artifact representation approach and the introduction of the Synthetic-Pop dataset. 

\textbf{Limitations and Future Work.} While the proposed method demonstrates robust performance, its effectiveness against compression-altered synthetic images has yet to be fully explored. Future work will focus on improving detection of compression-altered images and extending the approach to address challenges posed by emerging forms of synthetic media.

\newpage
\section*{Acknowledgements}

This work was supported by the National Key R\&D Program of China (2022YFC3303603), the National Natural Science Foundation of China (NSFC, 62377028), the Guangdong Basic and Applied Basic Research Foundation (2023B1515120064), the Guangzhou Science and Technology Planning Project (Nansha District: 2023ZD001), the Guangzhou Development District International Cooperation Project (Grant No. 2023GH01), and the Fundamental Research Funds for the Central Universities (21625102).

\bibliographystyle{plainnat}
\bibliography{main}

\newpage
\section*{Appendix}
\appendix
 
\section{Details of Synthetic-Pop}
\label{sec: details of dataset}

\begin{figure*}[htbp]
    \centering

    \newcommand{\alignedcaption}[3]{%
        \parbox[b][4.5em][c]{0.9\linewidth}{%
            \centering\small
            \begin{tabular}[t]{@{}c@{}}
                #1 \\[0.3em]
                #2 \\[0.3em]
                #3 \\[0.3em]
            \end{tabular}
        }%
    }

    \begin{subfigure}{0.32\linewidth}
        \centering
        \includegraphics[width=\linewidth]{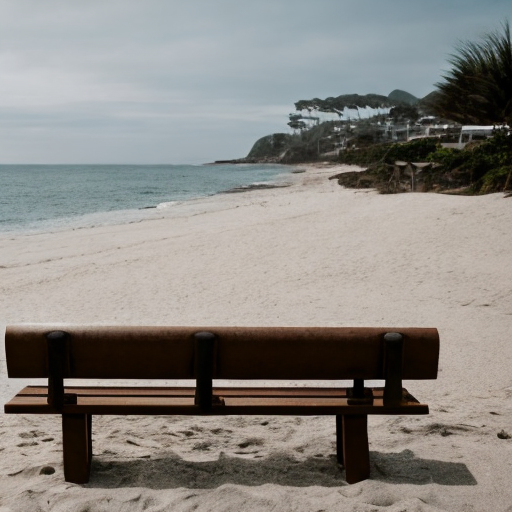}
        \alignedcaption
            {Openjourney: A bench sitting}
            {on the beach near the ocean.}
            {}
    \end{subfigure}
    \begin{subfigure}{0.32\linewidth}
        \centering
        \includegraphics[width=\linewidth]{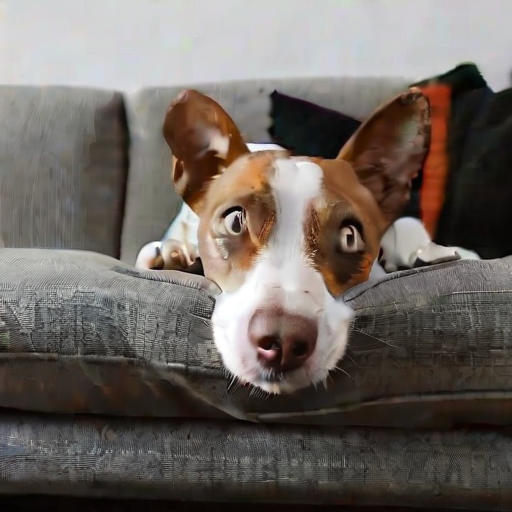}
        \alignedcaption
            {Proteus-0.3: A brown white}
            {and black dog is laying on a}
            {gray couch.}
    \end{subfigure}
    \begin{subfigure}{0.32\linewidth}
        \centering
        \includegraphics[width=\linewidth]{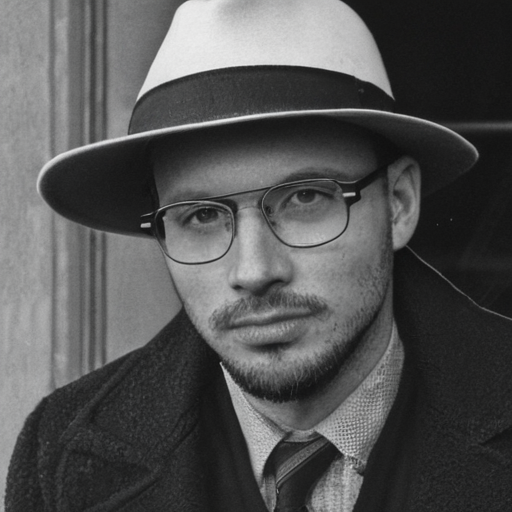}
        \alignedcaption
            {RealVisXL-4.0: A man that has}
            {glasses and a hat.}
            {}
    \end{subfigure}

    \begin{subfigure}{0.32\linewidth}
        \centering
        \includegraphics[width=\linewidth]{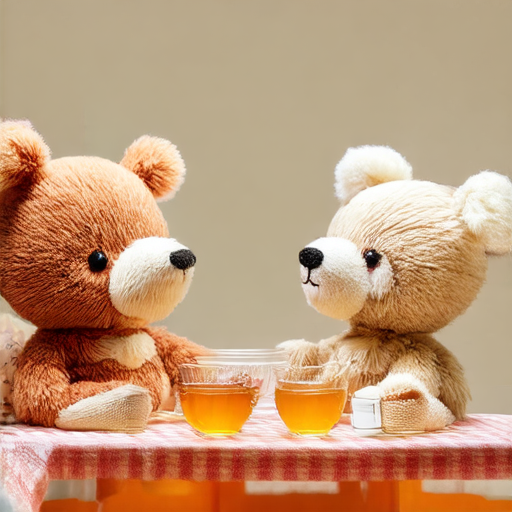}
        \alignedcaption
            {SD-3.5-Medium: Two stuffed}
            {animals sit at a table with honey.}
            {}
    \end{subfigure}
    \begin{subfigure}{0.32\linewidth}
        \centering
        \includegraphics[width=\linewidth]{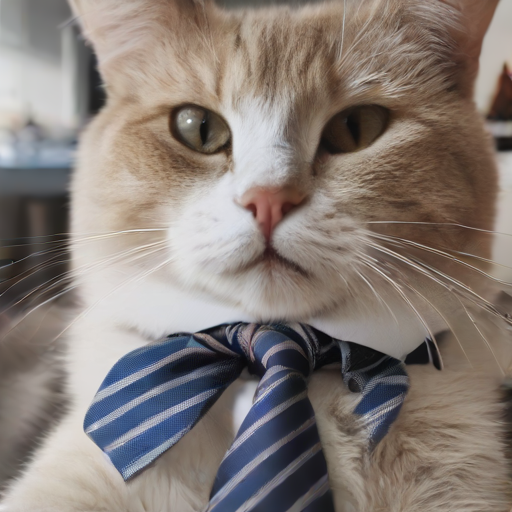}
        \alignedcaption
            {SDXL-Turbo: Cat sitting up}
            {with a fake tie around its neck.}
            {}
    \end{subfigure}
    \begin{subfigure}{0.32\linewidth}
        \centering
        \includegraphics[width=\linewidth]{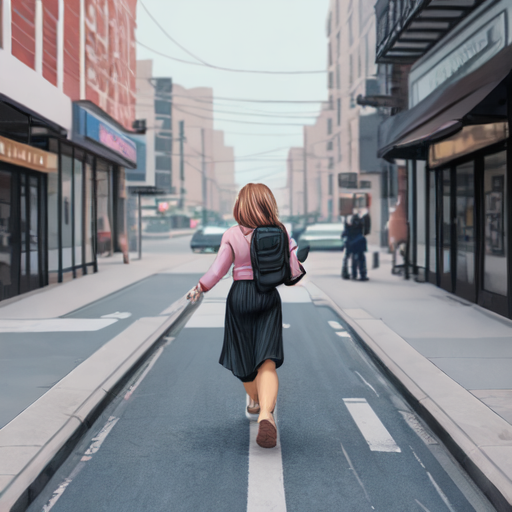}
        \alignedcaption
            {YiffyMix: A woman walking}
            {down a street talking on a}
            {cell phone.}
    \end{subfigure}

    \caption{Examples of images generated by different models along with their corresponding text prompts. Each subfigure presents an image produced by a specific model, where the format “Model: Prompt” denotes the generating model and its input description.}
    \label{fig:syn pop vis}
\end{figure*}

To evaluate the practicality of our method across mainstream generated models, we construct a benchmark named \textbf{Synthetic-Pop}, which includes six widely used models: Openjourney~\cite{Openjourney}, Proteus-0.3~\cite{proteus_v03}, RealVisXL-4.0~\cite{RealVisXL_V4}, SD-3.5-Medium~\cite{esser2024sd35}, SDXL-Turbo~\cite{podell2024sdxl}, and YiffyMix~\cite{YiffyMix}. Images are generated using 5,000 captions randomly sampled from COCO~\cite{lin2014microsoft}, following the inference configurations recommended by each model. Representative examples are shown in Figure~\ref{fig:syn pop vis}.

\section{Visualization Analysis}
\label{sec:  visualization analysis}

\subsection{Success Case Analysis}

\begin{figure*}[t]
    \centering
    \raisebox{6.0\height}{\parbox[c][0em][c]{0.1\linewidth}{\centering \textbf{Real}}}
    \begin{subfigure}{0.17\linewidth}
        \includegraphics[width=\linewidth]{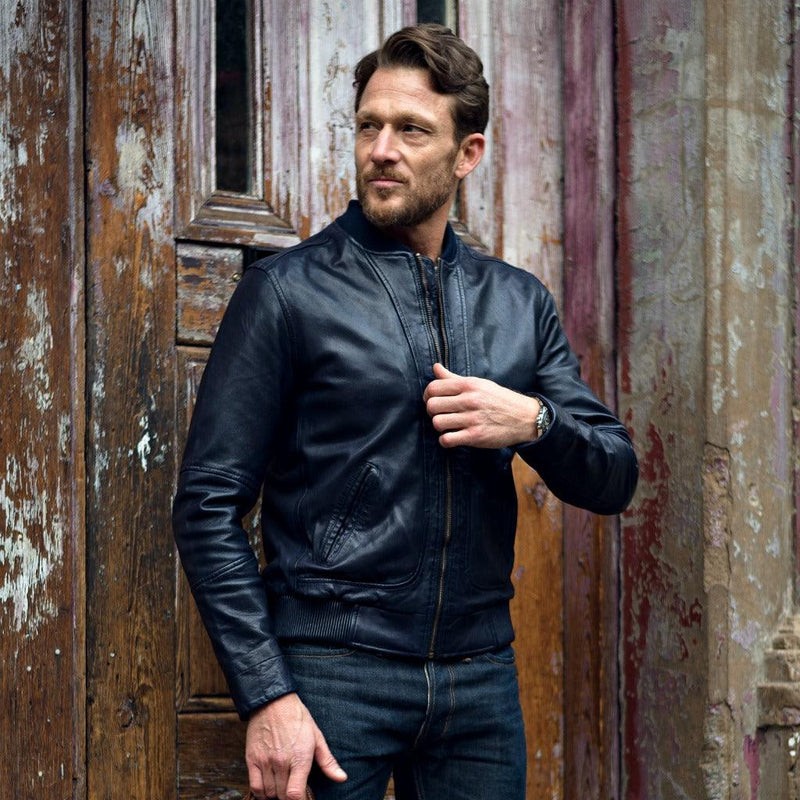}
    \end{subfigure}
    \begin{subfigure}{0.17\linewidth}
        \includegraphics[width=\linewidth]{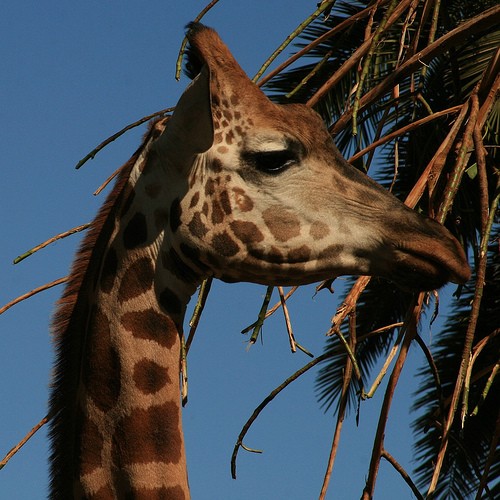}
    \end{subfigure}
    \begin{subfigure}{0.17\linewidth}
        \includegraphics[width=\linewidth]{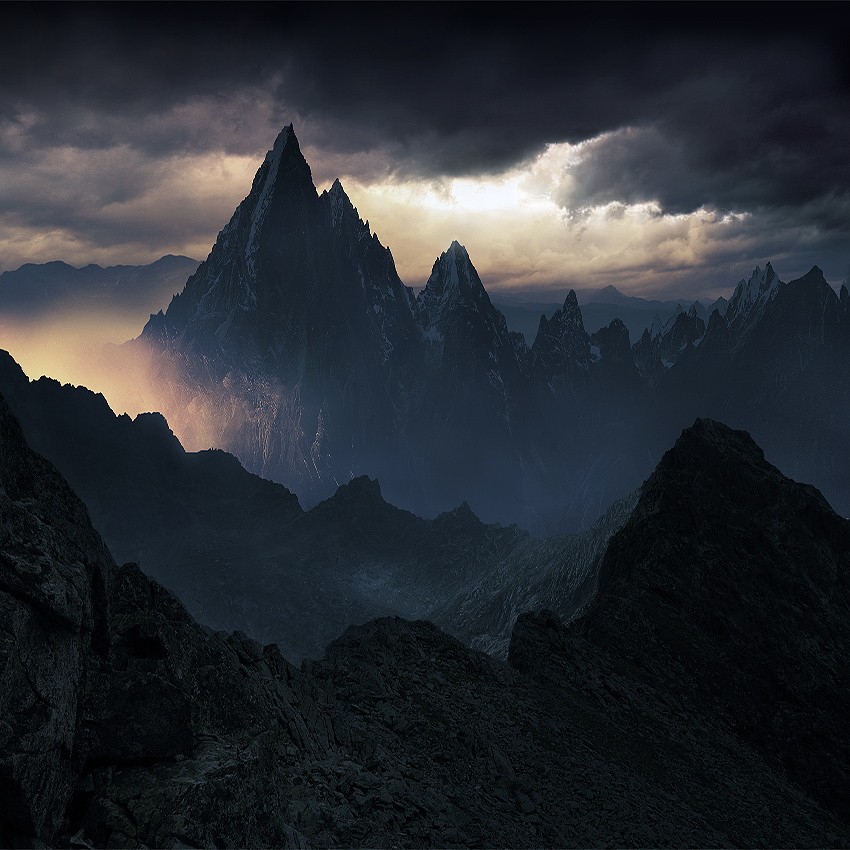}
    \end{subfigure}
    \begin{subfigure}{0.17\linewidth}
        \includegraphics[width=\linewidth]{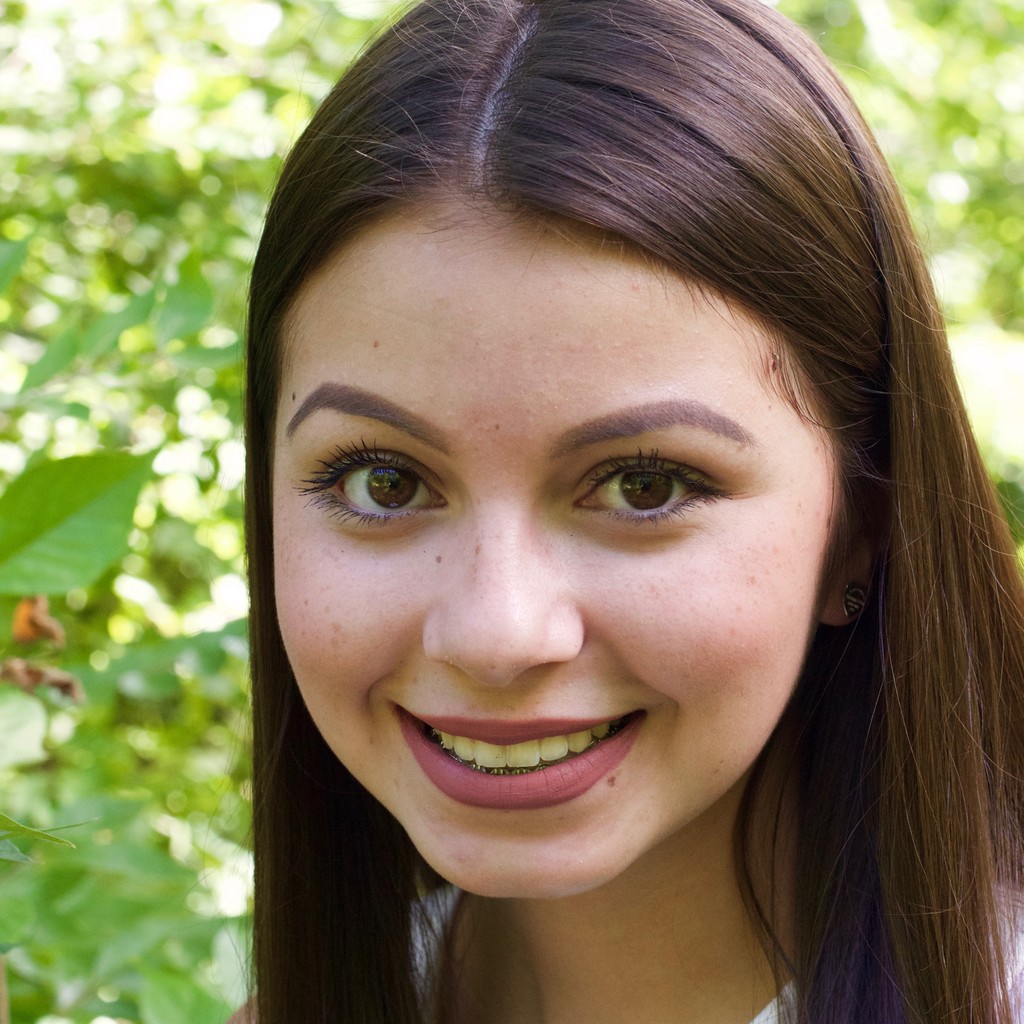}
    \end{subfigure}
    \begin{subfigure}{0.17\linewidth}
        \includegraphics[width=\linewidth]{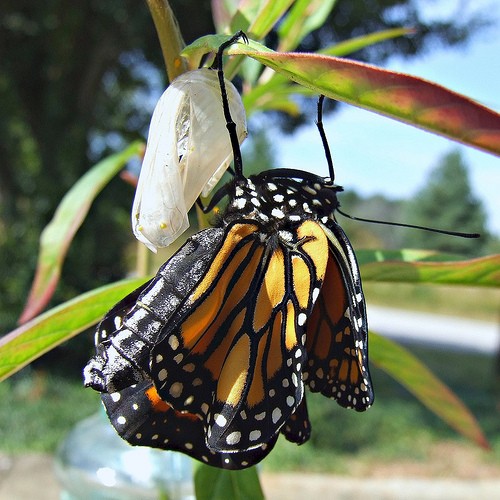}
    \end{subfigure}
    
    \raisebox{6.0\height}{\parbox[c][0em][c]{0.1\linewidth}{\centering \textbf{LPD}}}
    \begin{subfigure}{0.17\linewidth}
        \includegraphics[width=\linewidth]{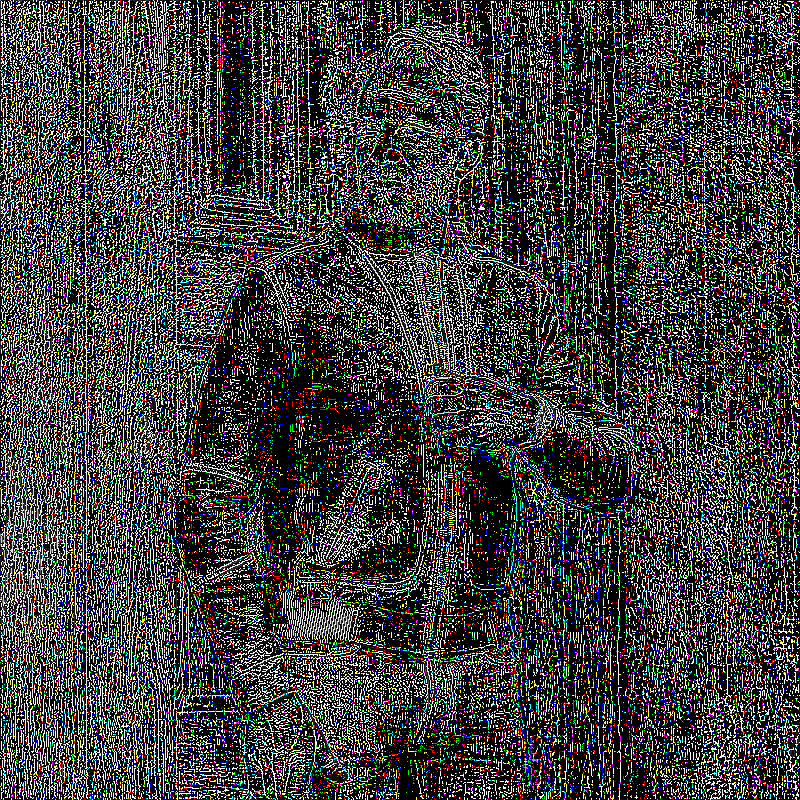}
    \end{subfigure}
    \begin{subfigure}{0.17\linewidth}
        \includegraphics[width=\linewidth]{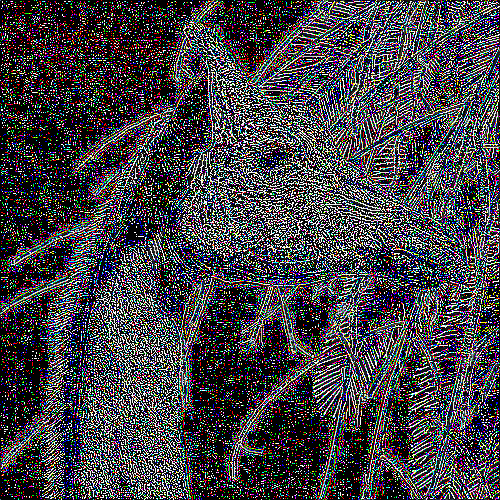}
    \end{subfigure}
    \begin{subfigure}{0.17\linewidth}
        \includegraphics[width=\linewidth]{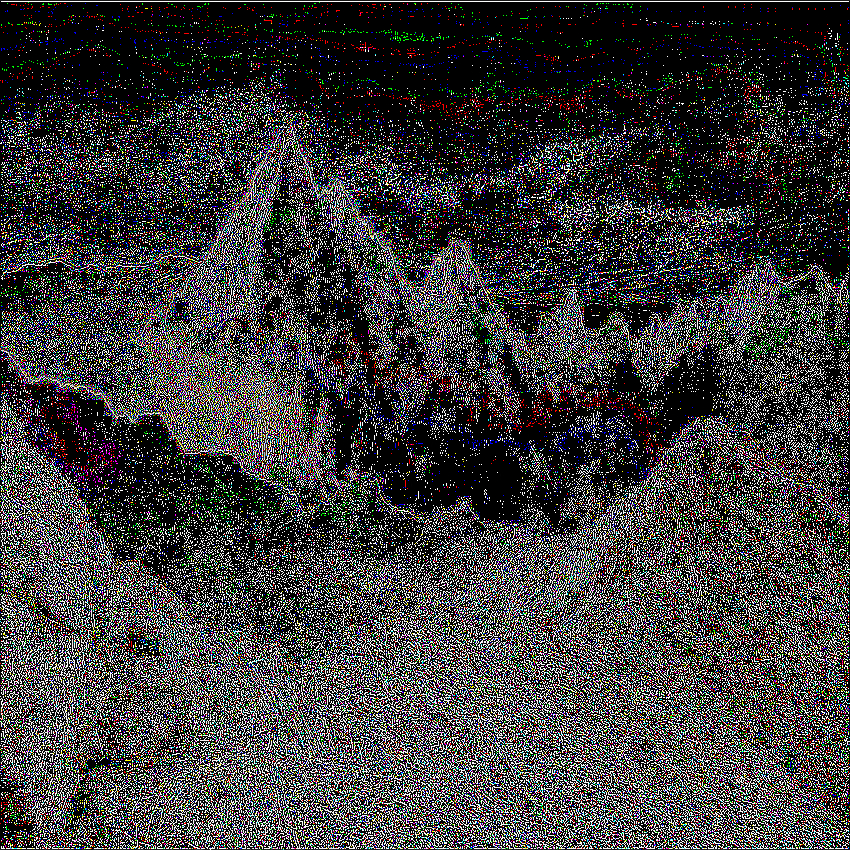}
    \end{subfigure}
    \begin{subfigure}{0.17\linewidth}
        \includegraphics[width=\linewidth]{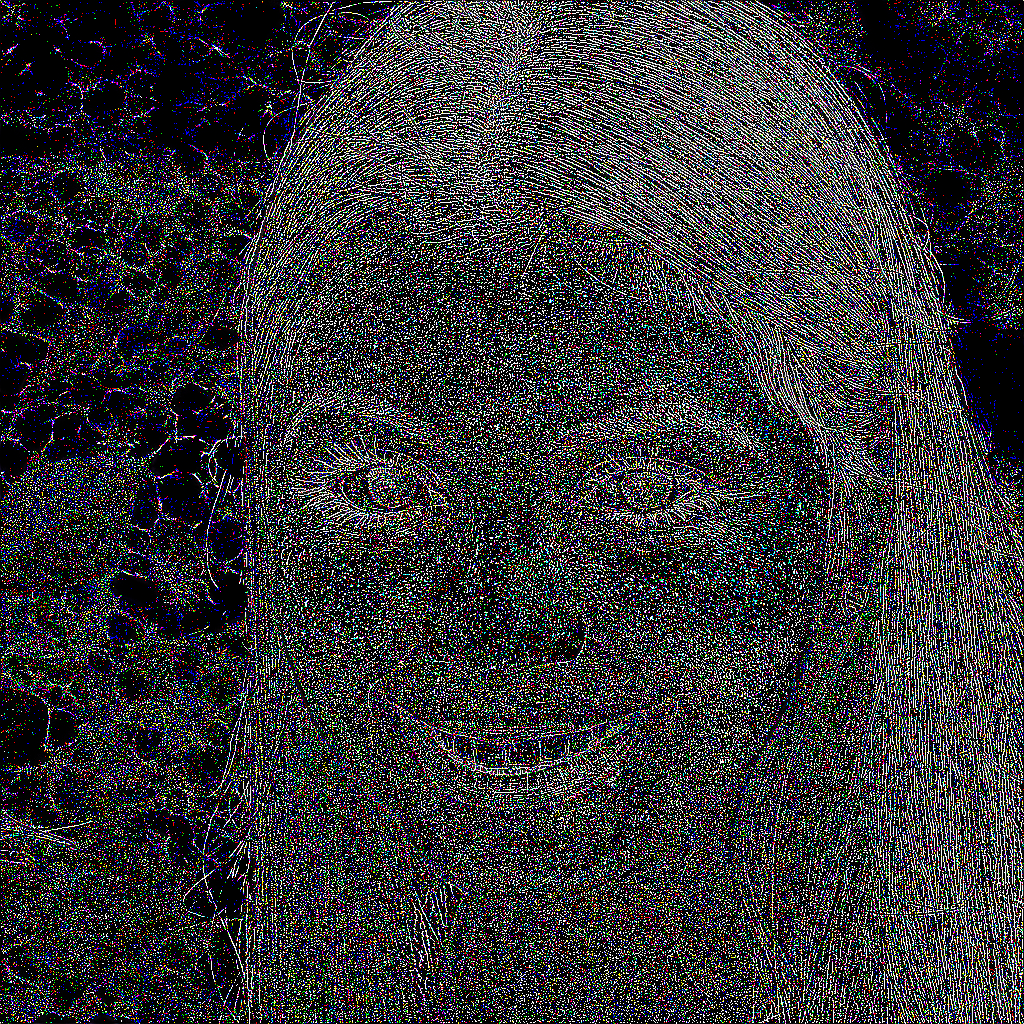}
    \end{subfigure}
    \begin{subfigure}{0.17\linewidth}
        \includegraphics[width=\linewidth]{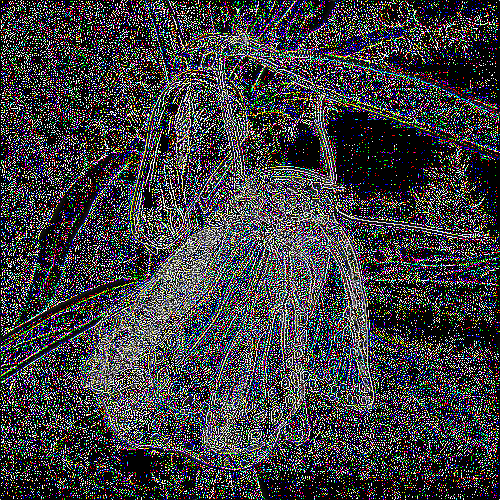}
    \end{subfigure}

    \raisebox{6.0\height}{\parbox[c][0em][c]{0.1\linewidth}{\centering \textbf{CAM}}}
    \begin{subfigure}{0.17\linewidth}
        \includegraphics[width=\linewidth]{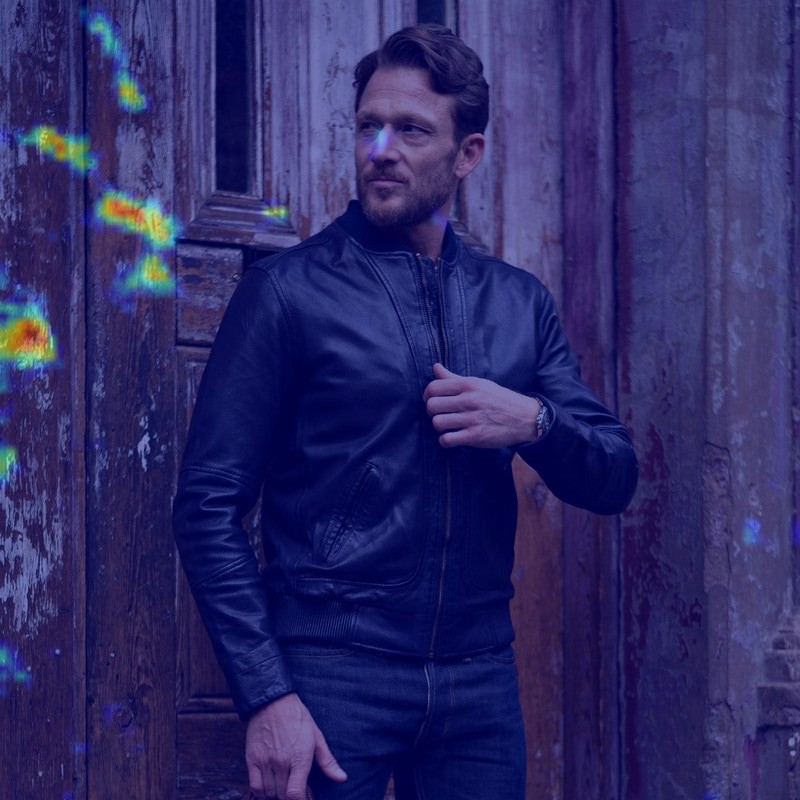}
    \end{subfigure}
    \begin{subfigure}{0.17\linewidth}
        \includegraphics[width=\linewidth]{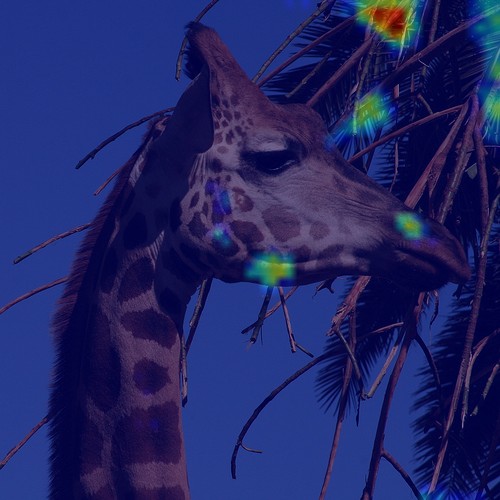}
    \end{subfigure}
    \begin{subfigure}{0.17\linewidth}
        \includegraphics[width=\linewidth]{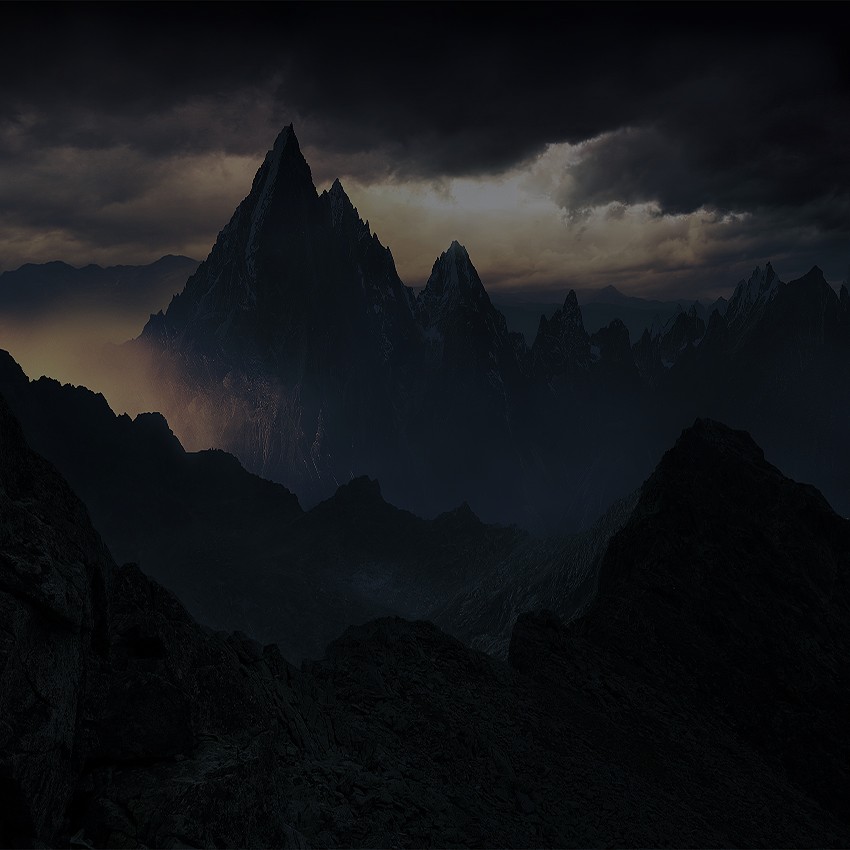}
    \end{subfigure}
    \begin{subfigure}{0.17\linewidth}
        \includegraphics[width=\linewidth]{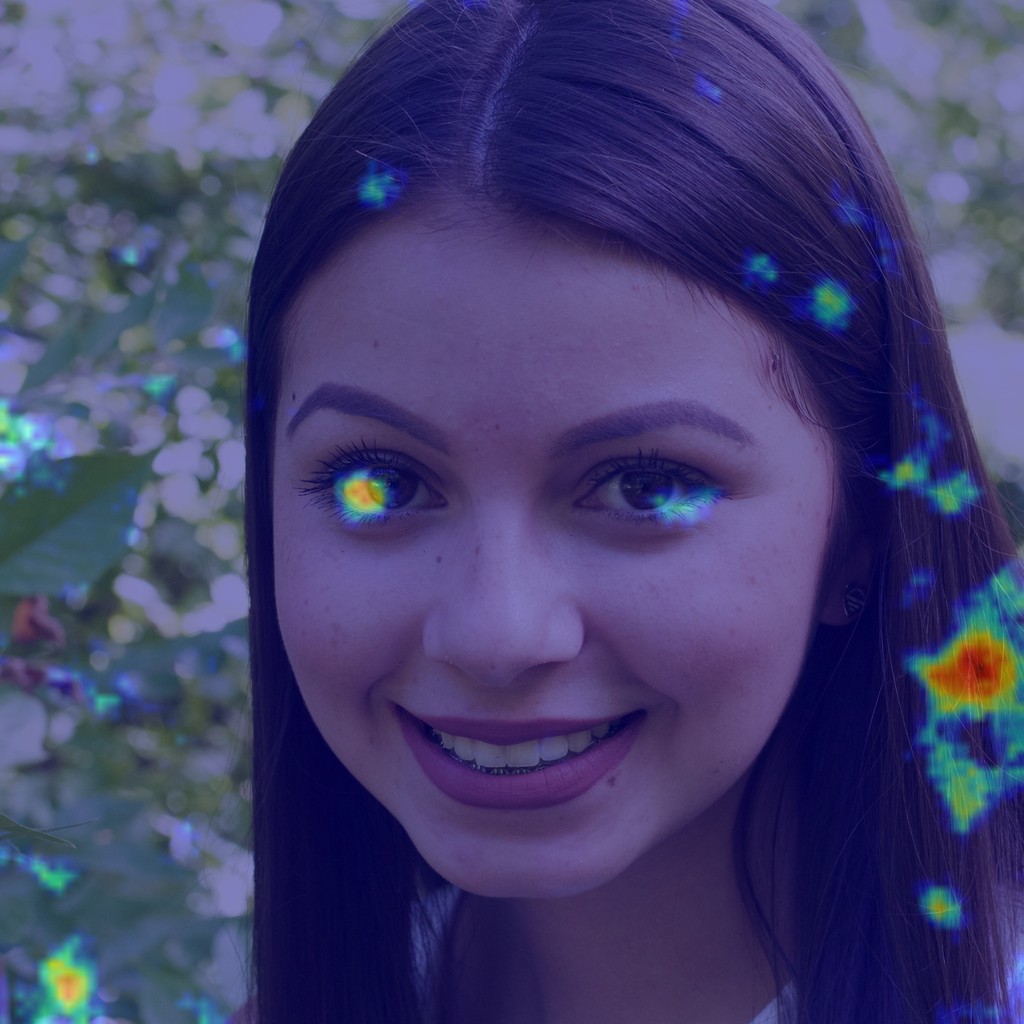}
    \end{subfigure}
    \begin{subfigure}{0.17\linewidth}
        \includegraphics[width=\linewidth]{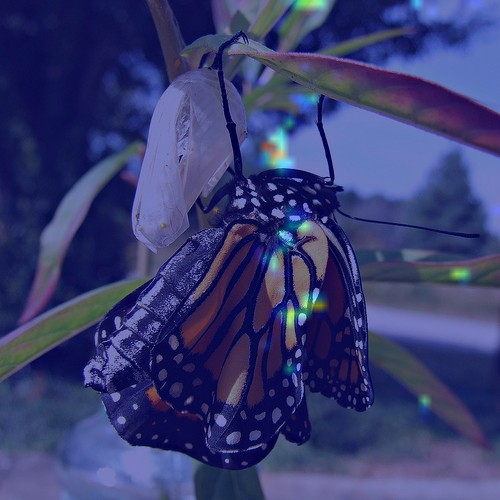}
    \end{subfigure}

    \raisebox{6.0\height}{\parbox[c][0em][c]{0.1\linewidth}{\centering \textbf{Fake}}}
    \begin{subfigure}{0.17\linewidth}
        \includegraphics[width=\linewidth]{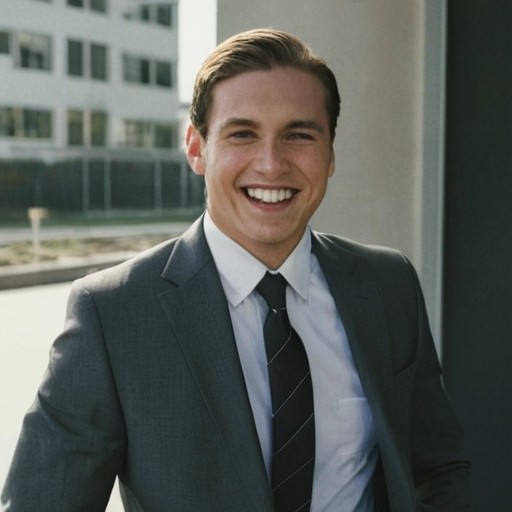}
    \end{subfigure}
    \begin{subfigure}{0.17\linewidth}
        \includegraphics[width=\linewidth]{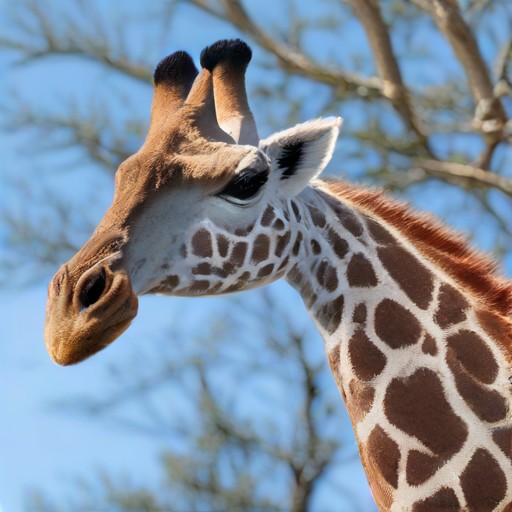}
    \end{subfigure}
    \begin{subfigure}{0.17\linewidth}
        \includegraphics[width=\linewidth]{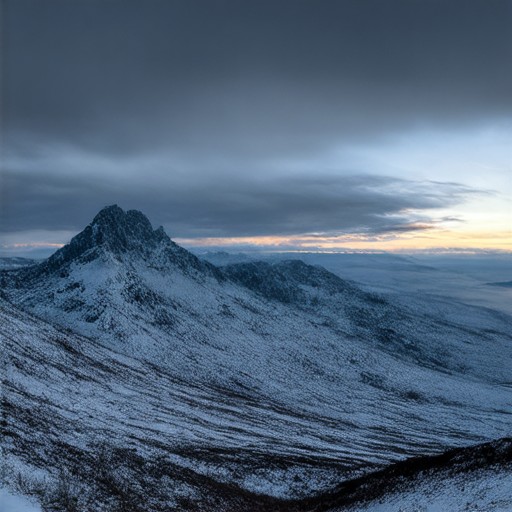}
    \end{subfigure}
    \begin{subfigure}{0.17\linewidth}
        \includegraphics[width=\linewidth]{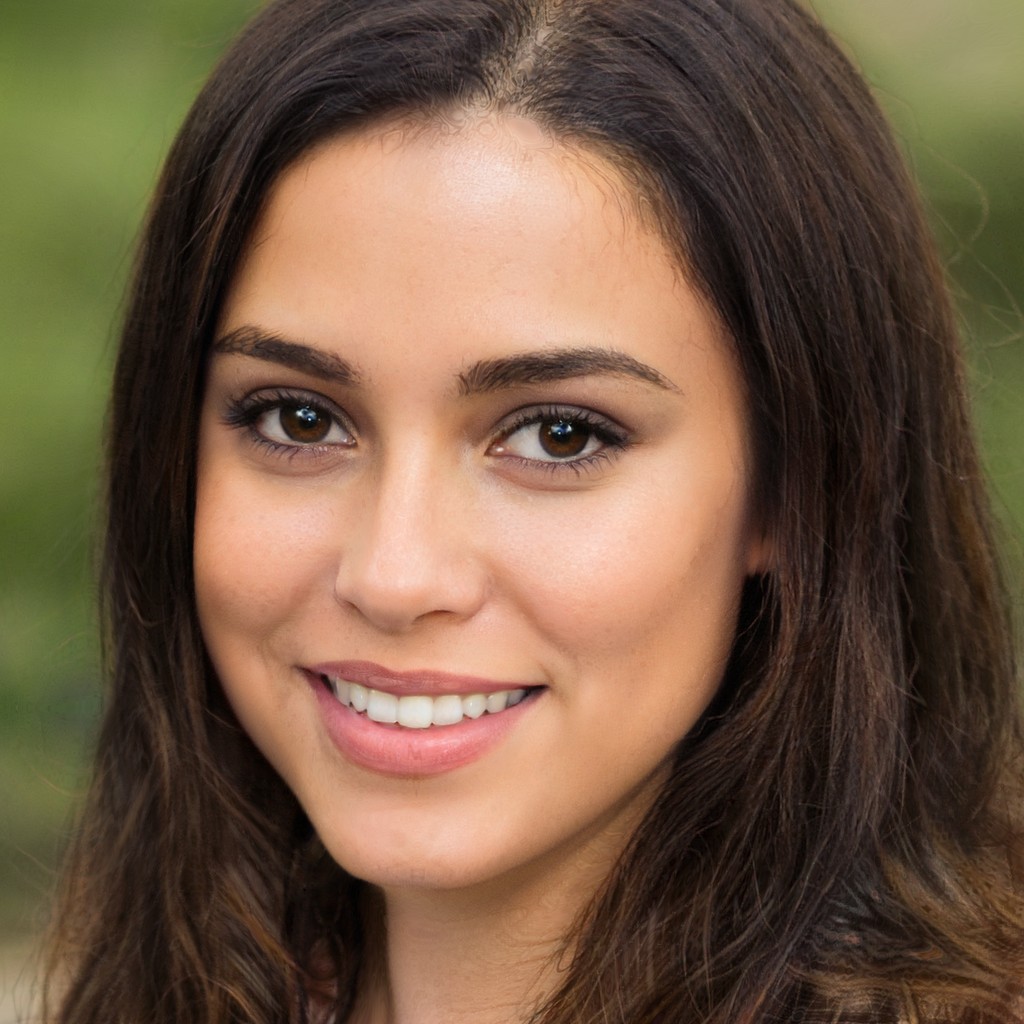}
    \end{subfigure}
    \begin{subfigure}{0.17\linewidth}
        \includegraphics[width=\linewidth]{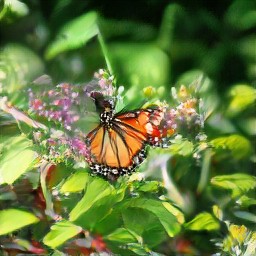}
    \end{subfigure}

    \raisebox{6.0\height}{\parbox[c][0em][c]{0.1\linewidth}{\centering \textbf{LPD}}}
    \begin{subfigure}{0.17\linewidth}
        \includegraphics[width=\linewidth]{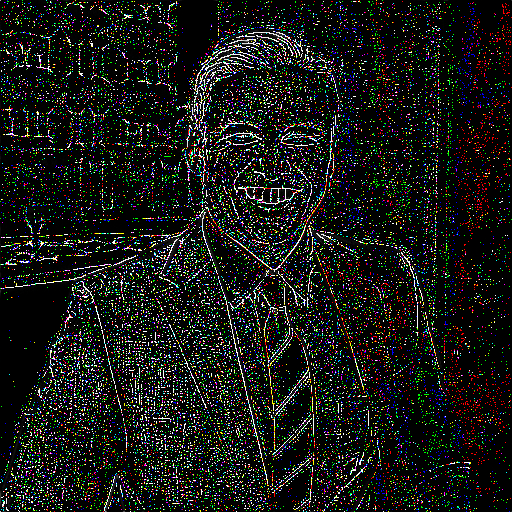}
    \end{subfigure}
    \begin{subfigure}{0.17\linewidth}
        \includegraphics[width=\linewidth]{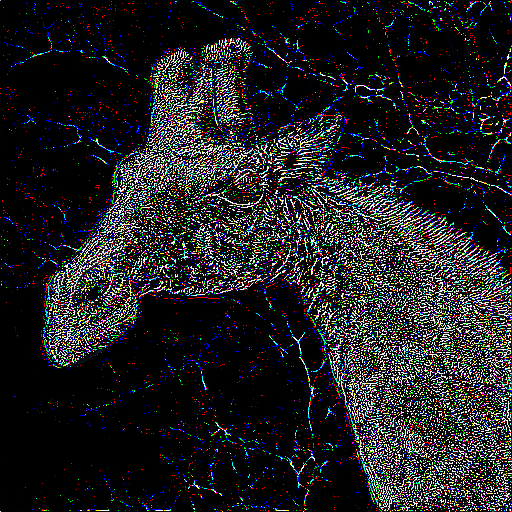}
    \end{subfigure}
    \begin{subfigure}{0.17\linewidth}
        \includegraphics[width=\linewidth]{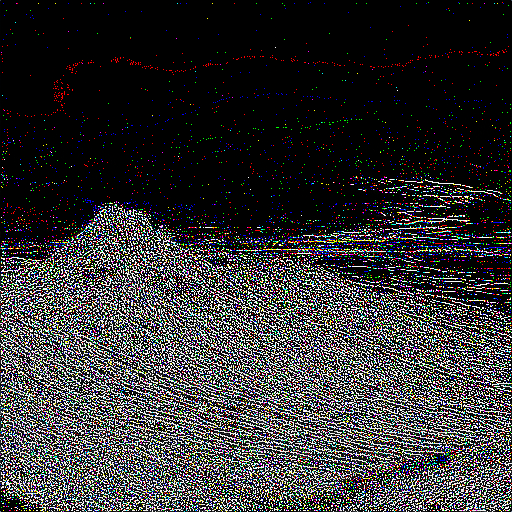}
    \end{subfigure}
    \begin{subfigure}{0.17\linewidth}
        \includegraphics[width=\linewidth]{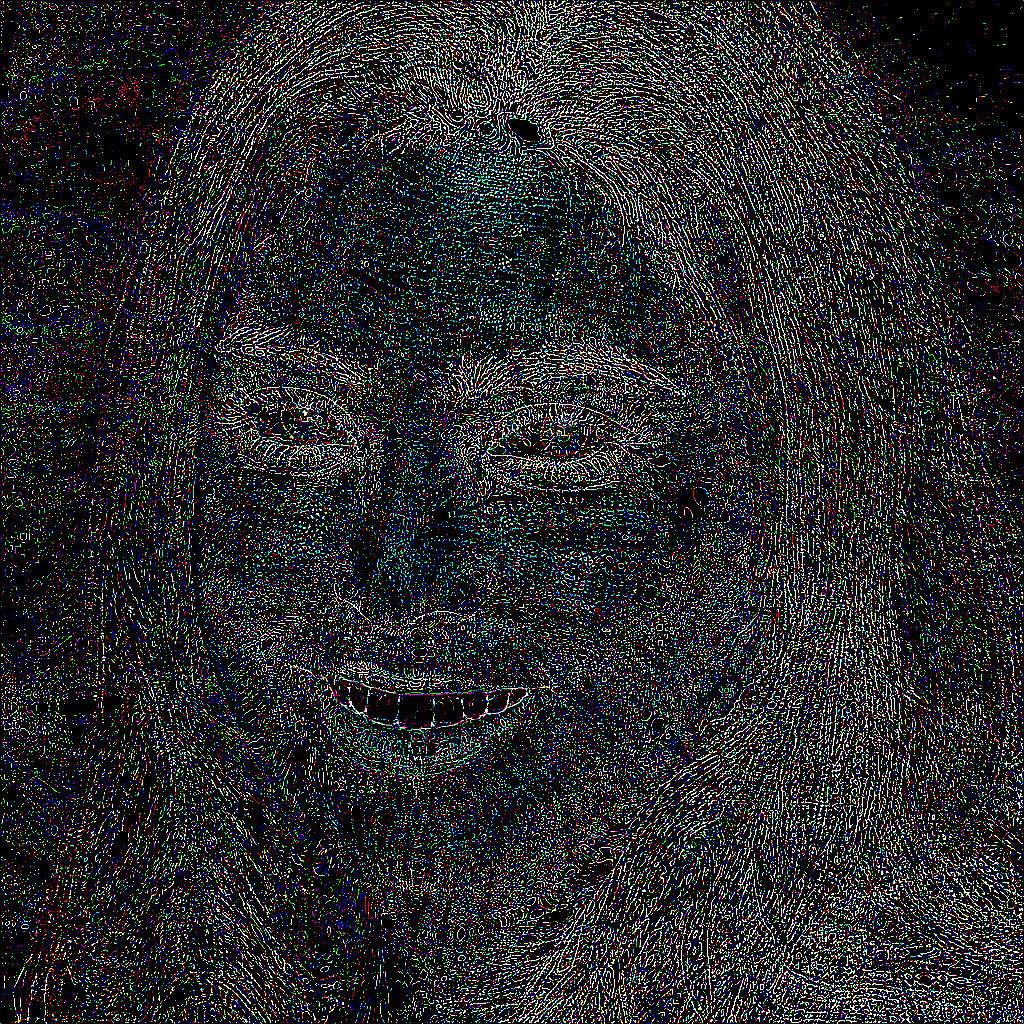}
    \end{subfigure}
    \begin{subfigure}{0.17\linewidth}
        \includegraphics[width=\linewidth]{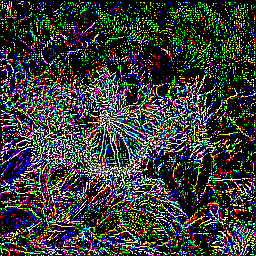}
    \end{subfigure}

    \raisebox{6.0\height}{\parbox[c][0em][c]{0.1\linewidth}{\centering \textbf{CAM}}}
    \begin{subfigure}{0.17\linewidth}
        \includegraphics[width=\linewidth]{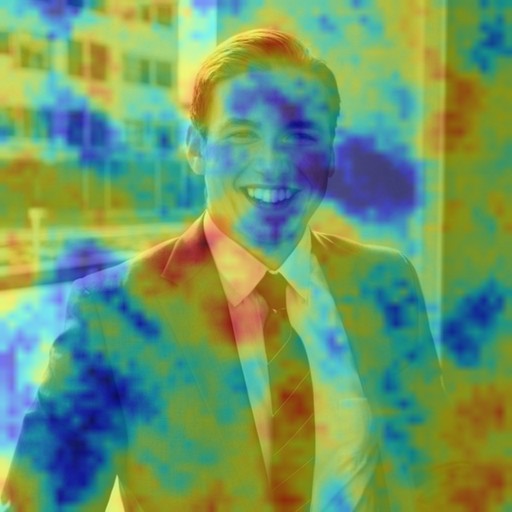}
    \end{subfigure}
    \begin{subfigure}{0.17\linewidth}
        \includegraphics[width=\linewidth]{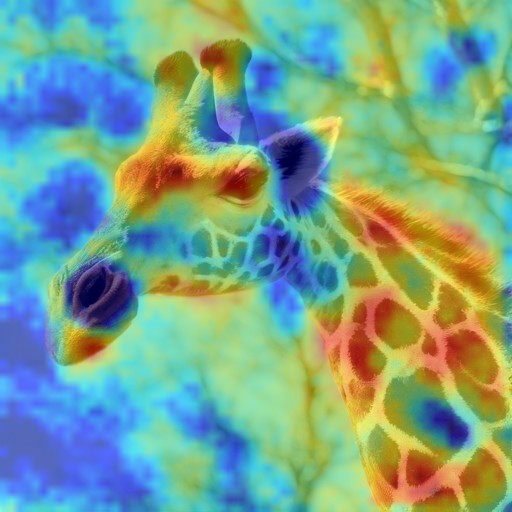}
    \end{subfigure}
    \begin{subfigure}{0.17\linewidth}
        \includegraphics[width=\linewidth]{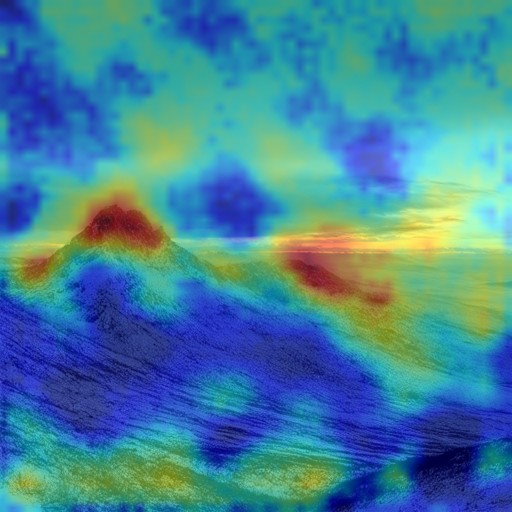}
    \end{subfigure}
    \begin{subfigure}{0.17\linewidth}
        \includegraphics[width=\linewidth]{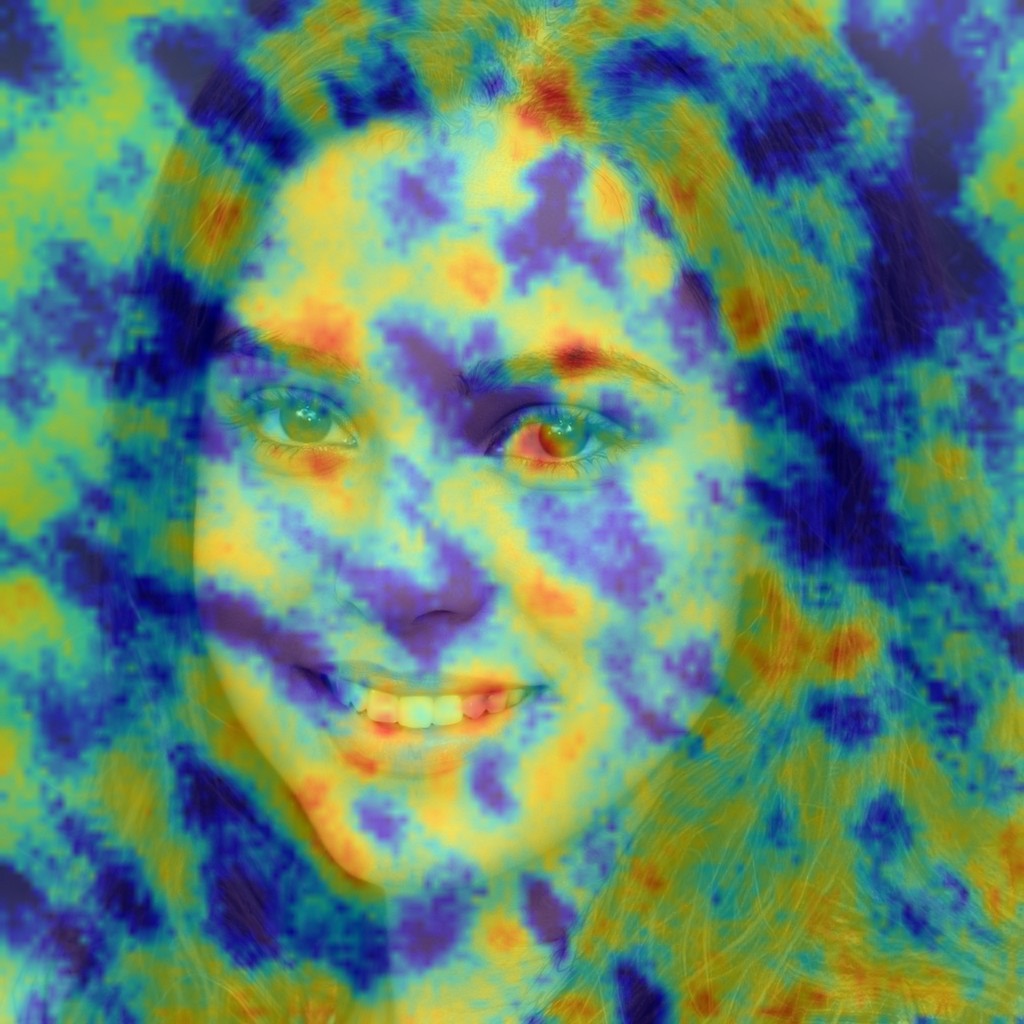}
    \end{subfigure}
    \begin{subfigure}{0.17\linewidth}
        \includegraphics[width=\linewidth]{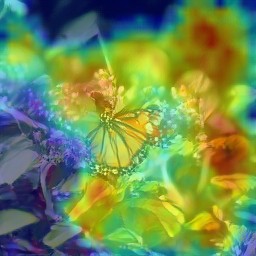}
    \end{subfigure}

    \caption{LPD and Grad-CAM visualizations of real and fake images.}
    \label{fig:success sample}
\end{figure*}

The effectiveness of LPD in distinguishing synthetic from real images stems from its ability to exploit intrinsic discrepancies in their local statistical structures. First, LPD captures subtle but systematic deviations introduced during the generative process. Although synthetic images may appear perceptually indistinguishable from real images, their micro-texture distributions and noise characteristics deviate from the stochastic sensor noise inherent to real image acquisition. As illustrated in the second and fifth rows of Figure~\ref{fig:success sample}, real images retain structured yet naturally irregular noise patterns, whereas synthetic counterparts exhibit overly smooth regions or artificial regularities—a direct consequence of the generative model’s learned priors.

Second, LPD operates in a content-agnostic manner. It consistently reveals statistical inconsistencies across diverse semantic domains, including human portraits, wildlife, and landscapes. This invariance to semantic content highlights the signal-level nature of LPD, ensuring robustness against the rapid evolution of generative models that continue to enhance perceptual fidelity but still leave detectable low-level statistical artifacts.

Finally, the Grad-CAM visualizations in the third and sixth rows confirm that the model’s attention aligns closely with LPD activation regions. Rather than focusing on semantic objects, FerretNet concentrates on areas exhibiting statistical anomalies, enabling consistent detection across diverse image categories. This strong alignment between LPD and Grad-CAM underpins the discriminative strength of our approach: the network remains nearly silent on authentic images while responding sharply to synthetic ones. 

\subsection{Failure Case Analysis}

\begin{figure*}[t]
    \centering
    \begin{subfigure}{0.3\linewidth}
        \centering
        \includegraphics[width=\linewidth]{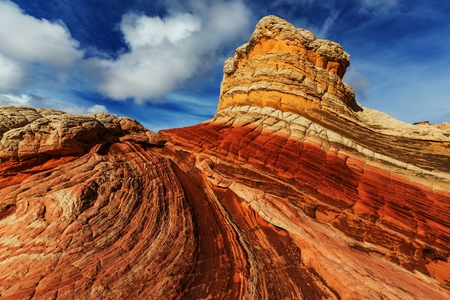}
    \end{subfigure}
    \begin{subfigure}{0.3\linewidth}
        \centering
        \includegraphics[width=\linewidth]{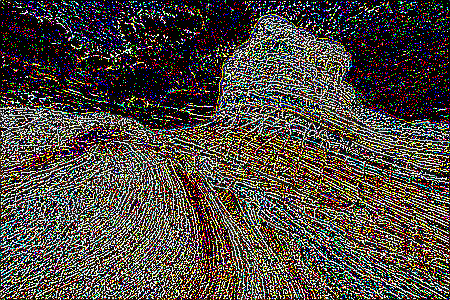}
    \end{subfigure}
    \begin{subfigure}{0.3\linewidth}
        \centering
        \includegraphics[width=\linewidth]{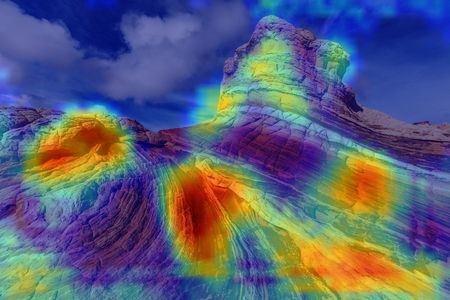}
    \end{subfigure}
    \\
    \begin{subfigure}{0.3\linewidth}
        \centering
        \includegraphics[width=\linewidth]{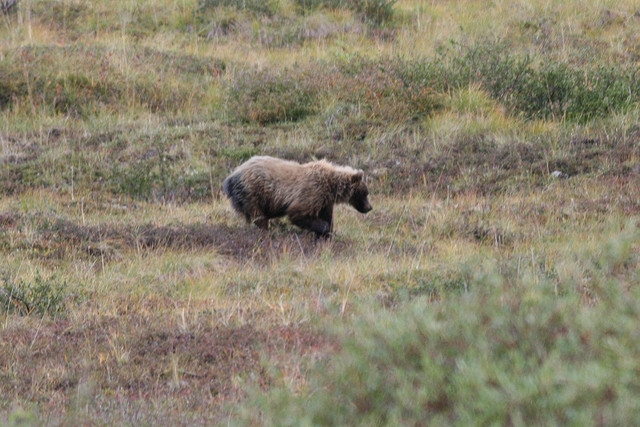}
    \end{subfigure}
     \begin{subfigure}{0.3\linewidth}
        \centering
        \includegraphics[width=\linewidth]{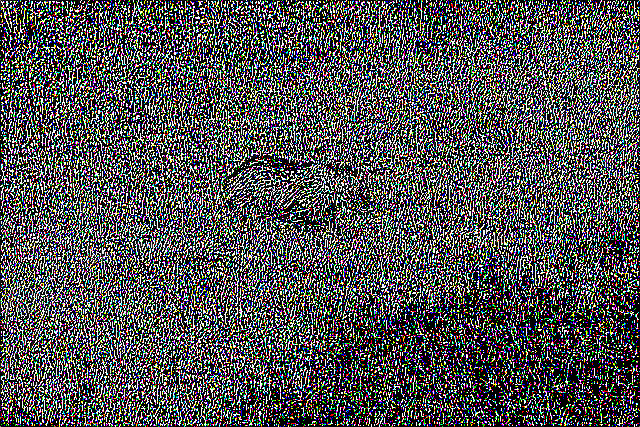}
    \end{subfigure}
    \begin{subfigure}{0.3\linewidth}
        \centering
        \includegraphics[width=\linewidth]{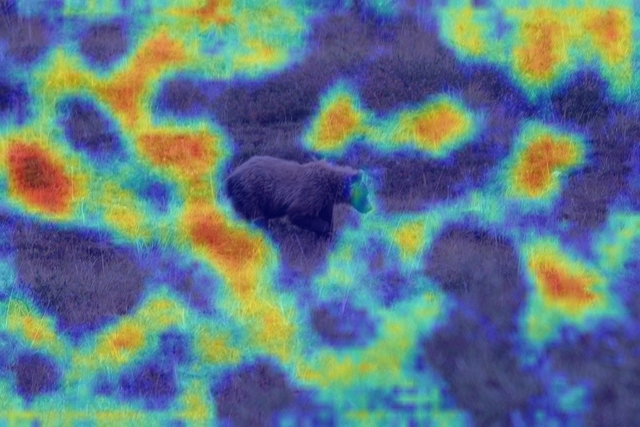}
    \end{subfigure}
    \\
    \caption{LPD and Grad-CAM visualizations of False Positive Case.}
    \label{fig:FP Samples}
\end{figure*}

\textbf{False Positive Case.} Figure~\ref{fig:FP Samples} illustrates representative real images are mistakenly classified as synthetic. Unlike the clear statistical distinctions observed in successful detections, these cases demonstrate that certain real-world textures can exhibit statistical patterns similar to those introduced by generative models.

In the first row, geological landscapes display highly regular and repetitive patterns, such as stratified rock formations with sharp color transitions and well-structured edges. These natural textures lead the LPD to capture statistical irregularities that resemble those typically found in generated content, particularly around boundary regions.

In the second row, the image of a bear on grassland demonstrates a similar issue. The network has learned to associate strong, structured statistical biases as primary indicators of artificial generation, resulting in misclassification. This reflects the model’s high sensitivity to subtle statistical cues, while also highlighting its limitations when encountering rare real-world scenes that are themselves highly structured “outliers.”

\begin{figure*}[t]
    \centering
    \begin{subfigure}{0.3\linewidth}
        \centering
        \includegraphics[width=\linewidth]{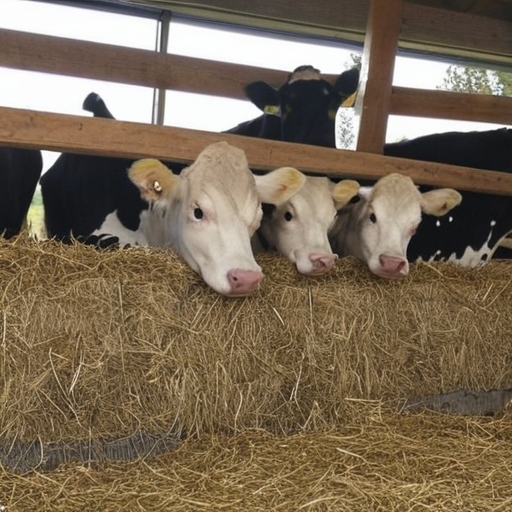}
    \end{subfigure}
    \begin{subfigure}{0.3\linewidth}
        \centering
        \includegraphics[width=\linewidth]{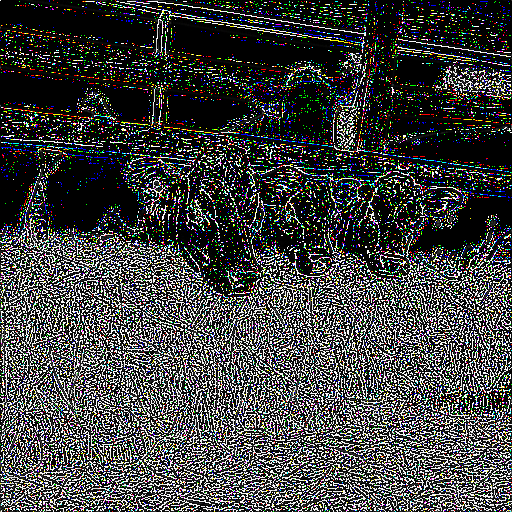}
    \end{subfigure}
    \begin{subfigure}{0.3\linewidth}
        \centering
        \includegraphics[width=\linewidth]{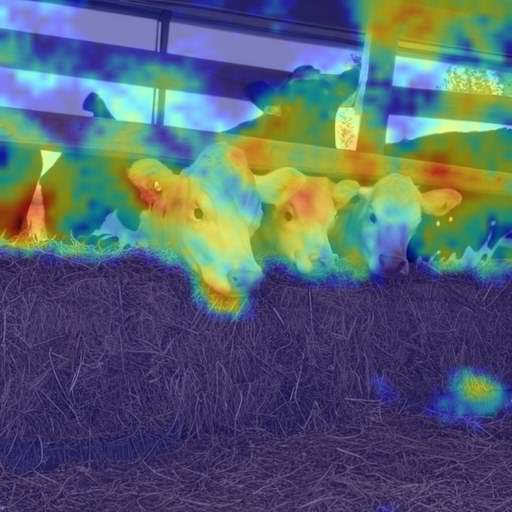}
    \end{subfigure}
    \\
    \begin{subfigure}{0.3\linewidth}
        \centering
        \includegraphics[width=\linewidth]{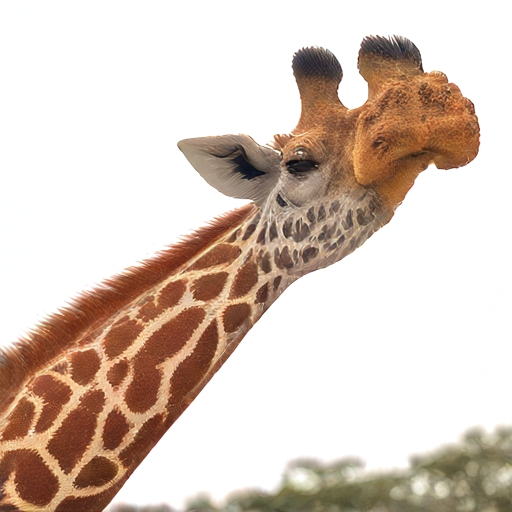}
    \end{subfigure}
     \begin{subfigure}{0.3\linewidth}
        \centering
        \includegraphics[width=\linewidth]{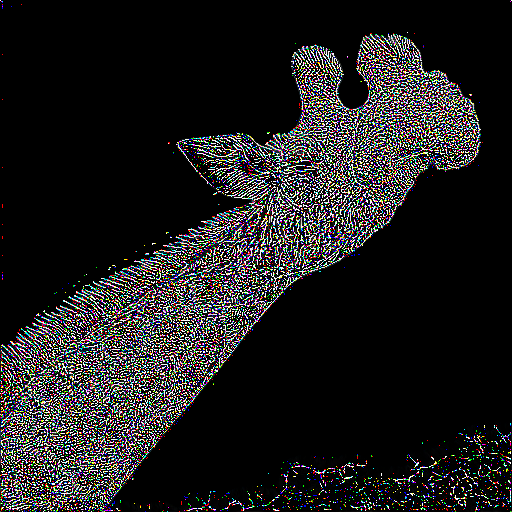}
    \end{subfigure}
    \begin{subfigure}{0.3\linewidth}
        \centering
        \includegraphics[width=\linewidth]{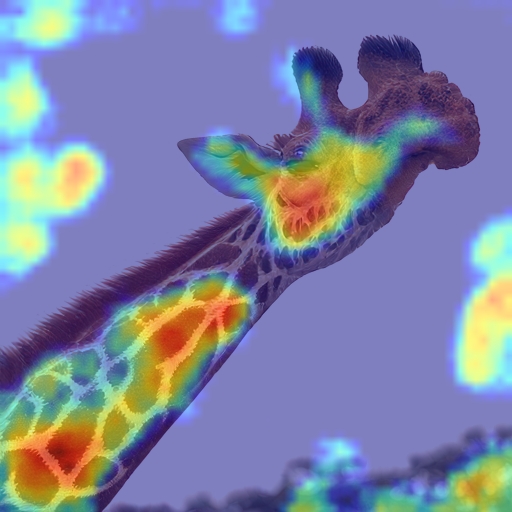}
    \end{subfigure}
    \\
    \caption{LPD and Grad-CAM visualizations of False Negative Case.}
    \label{fig:FN Samples}
\end{figure*}

\textbf{False Negative Case.} Figure~\ref{fig:FN Samples} illustrates representative synthetic images that were mistakenly classified as real. These cases reveal two distinct mechanisms by which AI-generated content can evade detection, highlighting limitations of LPD-based approaches.

In the synthetic cow image, the LPD map correctly highlights widespread, high-energy artifacts. Here, the failure is not in feature detection but in decision aggregation: the network sums all detected artifact signals to compute a final "fakeness" score. In this instance, despite numerous detected artifacts, their combined contribution did not exceed the threshold for a "fake" classification. This represents a rare boundary-case where a state-of-the-art generator produces content lying precisely on the “real” side of the learned decision boundary.

In contrast, the synthetic giraffe image exhibits a fundamentally different failure mode. Although Grad-CAM shows activation in textured regions, the overall signal is weak and insufficient for confident classification. A major factor is the clean, homogeneous sky background, which occupies a large portion of the image. Such smooth, featureless regions lack the local pattern variations necessary for LPD to detect pixel discontinuities or unnatural transitions. Consequently, the model receives insufficient discriminative signal, leading to misclassification.

\section{Additional Experiments}
\label{sec: add exp}
\subsection{Scaling FerretNet}
We scaled our model down (FerretNet-S) and up (FerretNet-L, ~20x parameters). As shown in Table~\ref{table:scaling}, making the model significantly larger results in a slight performance degradation. This confirms our hypothesis that for detecting low-level statistical artifacts, larger networks are prone to overfitting to training-specific patterns, which harms generalization. Our proposed FerretNet hits the "sweet spot" of being sufficiently expressive without the excess capacity.

\begin{table}[htbp]
\caption{Ablation on scaling FerretNet. Results are mean ACC/AP across all test sets.}
\centering
\begin{tabular}{l|ccc|c}
\toprule
Method & Channels & Blocks & Parameters & Mean ACC / AP \\
\midrule
FerretNet-S & (32, 64) & (2, 2) & 0.13 M & 93.1 / 97.6 \\
\textbf{FerretNet-B} & \textbf{(96, 192)} & \textbf{(2, 2)} & \textbf{1.06 M} & \textbf{97.1 / 99.6} \\
FerretNet-L & (96, 192, 384, 768) & (2, 2, 6, 2) & 21.51 M & 96.6 / 99.4 \\
\bottomrule
\end{tabular}
\label{table:scaling}
\end{table}

\subsection{Robustness to Common Post-Processing} 
We conducted a comprehensive robustness analysis on the ForenSynths~\cite{wang2020cnn} test set against JPEG compression, resizing, and rotation. For resizing, we used a stringent protocol with dynamic resolutions. As shown in Table~\ref{table:robustness}, FerretNet demonstrates strong robustness, particularly against rotation, where it significantly outperforms the heavyweight FatFormer~\cite{liu2024forgery}. This provides strong evidence that LPD features, being based on local, orientation-agnostic statistics, are inherently immune to geometric transformations. While heavy JPEG compression remains a challenge for all lightweight detectors, FerretNet performs competitively.

\begin{table}[htbp]
\caption{Robustness analysis on the ForenSynths test set against common post-processing attacks.}
\centering
\begin{tabular}{l|c|cc|cc|c}
\toprule
\multirow{2}{*}{Method} & \multirow{2}{*}{No Attack} & \multicolumn{2}{c|}{JPEG} & \multicolumn{2}{c|}{Resize} & Rotation \\
& & (Q=100) & (Q=75) & (S=0.75) & (S=1.25) & D=[-45°, 45°] \\
\midrule
FreqNet~\cite{tan2024frequency} & 91.5/98.5 & 50.5/66.6 & 50.1/51.8 & 65.2/85.8 & 64.9/82.8 & 79.9/91.6 \\
NPR~\cite{tan2024rethinking} & 92.5/96.1 & 55.0/59.3 & 50.0/49.1 & 83.9/84.9 & 78.9/81.8 & 86.7/90.7 \\
FatFormer~\cite{liu2024forgery} & 98.4/99.7 & 96.5/99.4 & 71.7/89.8 & --- & --- & 68.1/96.8 \\
\midrule
\textbf{FerretNet (Ours)} & \textbf{95.9/99.3} & \textbf{55.1/67.8} & \textbf{50.2/49.4} & \textbf{81.4/94.3} & \textbf{80.8/95.4} & \textbf{88.2/98.0} \\
\bottomrule
\end{tabular}
\label{table:robustness}
\end{table}

\end{document}